\documentclass[10pt,twocolumn,letterpaper]{article}

\usepackage[pagenumbers]{wacv}

\definecolor{wacvblue}{rgb}{0.21,0.49,0.74}
\usepackage{colortbl}
\usepackage{placeins}
\usepackage[pagebackref,breaklinks,colorlinks,allcolors=wacvblue]{hyperref}

\def\confName{arXiv}
\def\confYear{2026}

\title{CADENCE: Temporally Coherent Whole-Body Human Mesh Recovery\\
that Preserves Expressive Hand Motion}

\author{Keniel Peart \qquad Jagmohan Chauhan \qquad Indu P.\ Bodala \qquad Shelly Vishwakarma\\
University of Southampton, United Kingdom\\
{\small Project page: \url{https://keniel123.github.io/cadence} \quad Code: \url{https://github.com/keniel123/cadence}}
}

\begin{document}
\raggedbottom
\maketitle
\begin{abstract}
Whole-body human mesh recovery from monocular video requires accurate spatial
reconstruction and temporal consistency across the body, hands, and face.
Independent frame-wise estimation often introduces temporal jitter, while
smoothing can attenuate rapid expressive articulations. We present CADENCE,
a framework for temporally coherent whole-body SMPL-X recovery with fine-grained
hand reconstruction. A structured per-frame estimator and a keypoint-guided
hand pathway share a visual encoder. An offline temporal refiner combines
temporal and kinematic-tree attention to predict adaptive mixtures of symmetric
temporal filters for each joint and frame. On 3DPW, refinement reduces acceleration
error from 17.4 to 5.4\,mm/frame$^2$ while improving PA-MPJPE from 41.9 to
40.2\,mm. On SignAvatars/How2Sign sequences, CADENCE achieves the lowest hand
jitter among the evaluated configurations, with a hand-speed peak amplitude
ratio of 0.98 relative to pseudo-ground-truth motion, compared with 0.35 for a whole-body adaptation of
SmoothNet. On EHF, the whole-body estimator achieves a full-body mesh error of
25.7\,mm, and hand refinement achieves a hand joint error of 8.3\,mm.
These results support evaluating expressive recovery through spatial accuracy,
temporal stability, and preservation of rapid local articulation.
\end{abstract}

\section{Introduction}
\label{sec:intro}

Sign-language understanding, avatar animation, and human motion analysis need
3D motion of the body, hands, and face together~\cite{SMPL-X:2019,yu2024signavatars}.
SMPL-X~\cite{SMPL-X:2019} represents all three, and image-based estimators now
recover detailed SMPL-X meshes from single
frames~\cite{cai2023smplerx,lin2023one,sun2024aios,multi-hmr2024}. Applied
independently to video, these estimators exhibit temporal inconsistency. Frame-wise estimation errors manifest as high-frequency jitter in the limbs
and fingers and inconsistent global orientation estimates. Sign languages convey linguistic information through coordinated manual and
non-manual articulations, making accurate hand, face, and body motion essential.

Temporal mesh recovery is well studied for body-only
SMPL~\cite{kocabas2019vibe,choi2020beyond,WeiLin2022mpsnet,wham:cvpr:2024,wang2024tram},
but these models do not represent the hand, jaw, and expression parameters of
SMPL-X. The usual alternative for whole-body video is to run a per-frame SMPL-X
estimator and smooth its output. Conventional temporal refinement is primarily evaluated on body motion.
Expressive motion introduces heterogeneous temporal characteristics across
anatomical regions and activities. Fine-grained hand articulation often contains
higher-frequency motion than broader body movements. Signing frequently
alternates between sustained hand configurations and rapid articulatory
transitions. Temporal smoothing must accommodate these changes while reducing
estimation noise. Body-centric benchmarks do not capture the effect of temporal
regularisation on fine-grained hand articulation.

Whole-body video recovery introduces additional challenges in spatial resolution
and articulated motion modelling. Fingers occupy relatively few pixels in
whole-body images, motivating high-resolution crops and specialised
branches~\cite{Moon_2022_CVPRW_Hand4Whole,pymafx2023}. Temporal processing must
also account for the kinematic tree, since wrist rotation influences all
subsequent finger positions.

We present CADENCE, a whole-body SMPL-X system for monocular video. A per-frame
estimator built on a shared C-RADIOv4 encoder~\cite{ranzinger2026cradiov4techreport}
and a structured SMPL-X decoder produces each frame's estimate, and a
high-resolution hand pathway re-encodes hand crops with the same encoder
weights to refine finger pose. An offline temporal refiner then processes the
SMPL-X rotation sequence. Transformer attention over time and along the
kinematic tree predicts, for every joint and every frame, mixing weights over a
small bank of trainable symmetric temporal filters, initialised with responses
ranging from near-identity to stronger smoothing. The selection starts low-pass for body joints and near-identity for
hand, forearm, and jaw joints, so that body and hands are smoothed at different
rates, and it can change within a clip as a hand moves between held handshapes
and fast strokes.

On 3DPW, the refiner reduces acceleration error from 17.4 to
5.4\,mm/frame$^2$ and lowers PA-MPJPE from 41.9 to 40.2\,mm relative to the
per-frame input; five generic smoothers applied to the same predictions reach
6.2 to 7.7\,mm/frame$^2$. On SignAvatars/How2Sign sign-language sequences, where
the trade-off between stability and hand motion is visible, the refiner has the
lowest hand jitter among the compared smoothers and keeps a hand-speed peak
amplitude of 0.98, while a SmoothNet retrained on the full SMPL-X vector keeps
0.35. A single-rate version of the refiner, with one bandwidth shared by all
joints, reaches comparable jitter (1.50 against 1.29) but keeps a peak amplitude
of only 0.50.

Our contributions are threefold.
\begin{itemize}
  \item A kinematic-aware temporal refiner for SMPL-X rotation sequences that
  predicts joint- and frame-dependent mixtures of symmetric temporal filters
  to improve motion stability while preserving rapid local articulation
  (\cref{sec:temporal}), evaluated against generic smoothers and a single-rate
  variant (\cref{sec:temporal_results}).
  \item A high-resolution hand pathway that refines finger pose from hand crops
  with the shared encoder and no separate hand encoder (\cref{sec:hand}),
  reducing EHF hand P-MVE from 22.6 to 12.2\,mm (\cref{sec:hand_ablation}).
  \item An evaluation combining spatial accuracy, temporal stability, and
  hand-motion preservation on expressive whole-body and video benchmarks
  (\cref{sec:temporal_results,sec:ehf,sec:generalization}).
\end{itemize}

\section{Related Work}
\label{sec:related}

\paragraph{Single-image and whole-body human mesh recovery.}
Human mesh recovery was first developed mainly for body-only
SMPL~\cite{loper2015smpl}, with progress driven by iterative regression,
optimisation-in-the-loop training, part-guided attention, and camera-aware
conditioning~\cite{hmrKanazawa17,kolotouros2019spin,kocabas2021pare,li2022cliff,patel2024camerahmr}.
Transformer-based models further improved monocular recovery through stronger
global features and token-based reasoning~\cite{goel2023humans,dwivedi_cvpr2024_tokenhmr,sarandi2024nlf}.
For expressive recovery, SMPL-X~\cite{SMPL-X:2019} extends the body model with
articulated hands, face, jaw, expression, and shape. Early whole-body systems
such as ExPose, Hand4Whole, and PyMAF-X~\cite{choutas2020expose,Moon_2022_CVPRW_Hand4Whole,pymafx2023}
used separate crops or specialised branches for body, hands, and face, improving
local detail but increasing pipeline complexity. Recent unified methods,
including SMPLer-X~\cite{cai2023smplerx}, OSX~\cite{lin2023one},
AiOS~\cite{sun2024aios}, Multi-HMR~\cite{multi-hmr2024},
PromptHMR~\cite{wang2025prompthmr}, PEAR~\cite{wu2026pearpixelalignedexpressivehuman},
and SAM~3D~Body~\cite{sam3dbody}, process the person more holistically and
achieve strong single-image results.

SAM 3D Body uses a shared image encoder, transformer decoding, and hand
refinement for single-image reconstruction in the Momentum Human Rig
representation. CADENCE predicts SMPL-X and adds pose-space temporal refinement
to address consistency of expressive video motion.

\paragraph{Hand detail and multi-scale visual features.}
Fine hand articulation is difficult to recover from a full-body crop because
fingers occupy only a small image region. Multi-scale representations help
combine semantic context with local precision~\cite{lin2017feature,sun2019deep},
and pose-estimation work such as ViTPose~\cite{xu2022vitpose} shows earlier
transformer features retain useful localisation detail. Many whole-body
methods use high-resolution hand crops and hand-specific
encoders~\cite{Moon_2022_CVPRW_Hand4Whole,pymafx2023}, improving hand detail
but separating the hand estimate from global body context and adding system
complexity. CADENCE instead reuses its shared visual encoder on hand crops to predict
finger pose, retaining the body, camera, shape, and face predictions from the
whole-body estimate.

\paragraph{Video-based and temporal mesh recovery.}
Temporal mesh recovery has advanced mainly for body-only SMPL, using temporal
convolutions~\cite{kanazawa2019learning}, adversarial motion priors
~\cite{kocabas2019vibe}, past--future fusion~\cite{choi2020beyond}, per-joint
temporal modelling~\cite{WeiLin2022mpsnet}, and world- or camera-aware
trajectories~\cite{wham:cvpr:2024,wang2024tram}. Post-hoc methods such as
SmoothNet~\cite{zeng2022smoothnet} and MotionBERT~\cite{motionbert2022} can also
smooth existing predictions, while HTD-Refine~\cite{htdrefine} improves motion
through test-time optimisation. These approaches are mainly designed and
evaluated for body pose, whereas expressive SMPL-X video recovery must account
for hands, face, and body together.

DanceHMR~\cite{shen2026dancehmr} is the closest whole-body video method to
CADENCE: it recovers temporally coherent body and hand SMPL-X motion by
injecting hand evidence as a residual into a body-centric token before a
single shared temporal transformer, using frozen per-frame body and hand
feature extractors and a close-up-aware training strategy that synthesizes
truncated crops to handle vlog- and livestream-style upper-body footage.
CADENCE refines motion directly in SMPL-X rotation space using temporal and
kinematic-tree attention to select a symmetric filter mixture for every joint
and frame. This explicitly parameterises adaptive temporal bandwidth and allows
its effect on stability and hand-speed preservation to be evaluated directly
(\cref{sec:temporal_results}).

\section{CADENCE}
\label{sec:method}

\paragraph{Overview.}
Given monocular video frames $\mathbf{I}_1,\ldots,\mathbf{I}_T$ and per-frame person boxes,
CADENCE estimates a sequence of SMPL-X~\cite{SMPL-X:2019} parameters. SMPL-X
maps pose, shape, and facial expression parameters to a mesh
$\mathbf{V}\in\mathbb{R}^{10475\times3}$, while camera parameters determine
its image-space projection. At each frame, we predict a
341-dimensional parameter vector
\begin{equation}
\mathbf{p}_t=[
\underbrace{\boldsymbol{\theta}^{\mathrm{body}}_t}_{126},
\underbrace{\boldsymbol{\theta}^{\mathrm{lh}}_t}_{90},
\underbrace{\boldsymbol{\theta}^{\mathrm{rh}}_t}_{90},
\underbrace{\boldsymbol{\theta}^{\mathrm{jaw}}_t}_{6},
\underbrace{\boldsymbol{\psi}_t}_{10},
\underbrace{\boldsymbol{\theta}^{\mathrm{or}}_t}_{6},
\underbrace{\boldsymbol{\beta}_t}_{10},
\underbrace{\mathbf{c}_t}_{3}],
\label{eq:layout}
\end{equation}
where rotations use the continuous 6D representation~\cite{Zhou2018OnTC}.
The body pose excludes the root orientation, which is predicted separately as
$\boldsymbol{\theta}^{\mathrm{or}}_t$.

CADENCE has three parts. First, a per-frame whole-body estimator predicts an
initial SMPL-X estimate from each image. Second, a high-resolution hand pathway
reuses the same visual encoder on hand crops and predicts finger pose.
Third, an offline temporal refiner processes the per-frame pose sequence. The
hand branch changes only finger-pose parameters; wrist, body-pose, global
orientation, shape, expression, face, and camera parameters are retained from
the per-frame estimate.

\begin{figure*}[t]
\centering
\includegraphics[width=\linewidth]{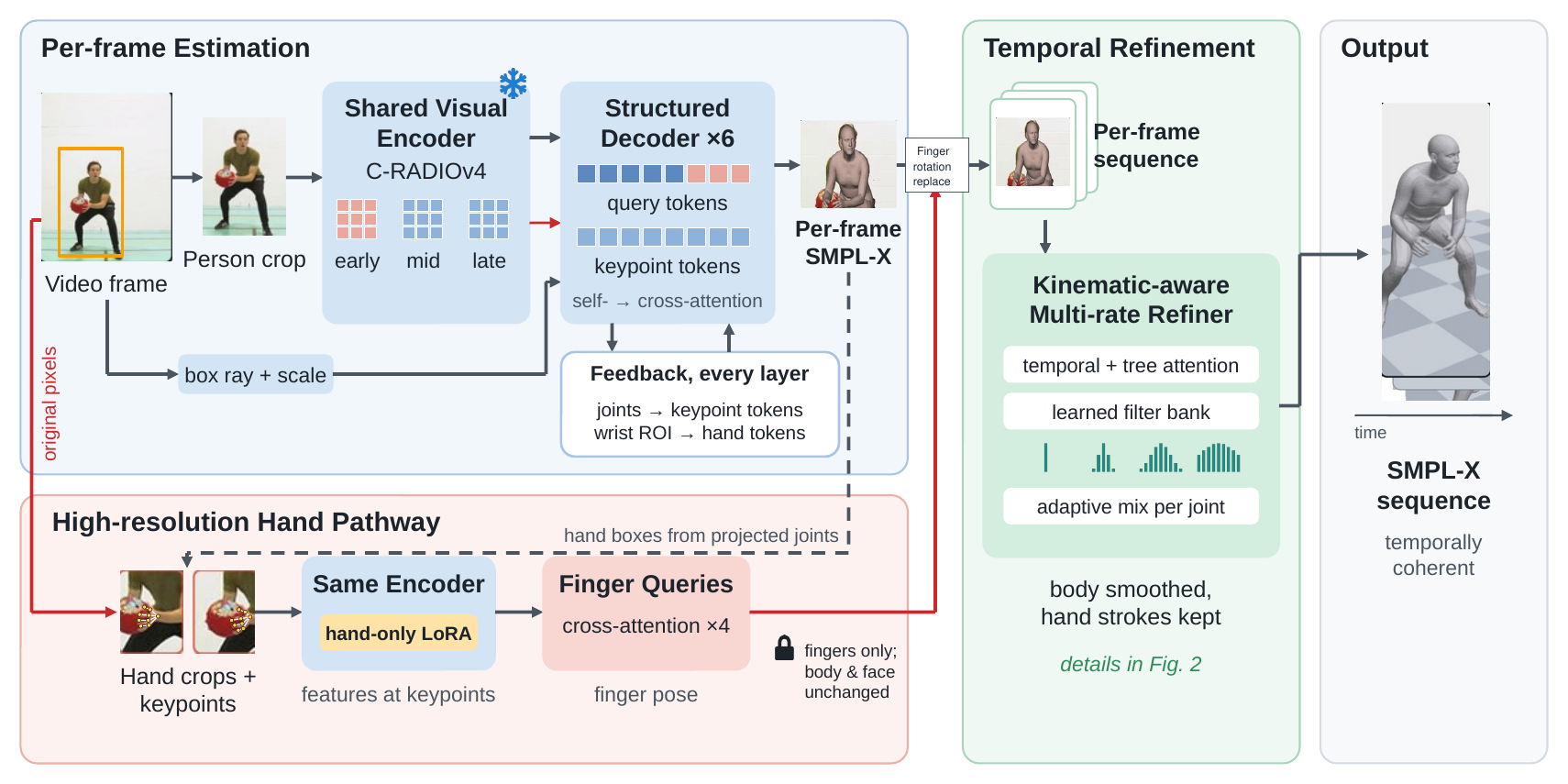}
\caption{CADENCE overview. The shared visual encoder and structured decoder
estimate per-frame SMPL-X parameters with iterative spatial feedback.
High-resolution hand crops reuse the encoder with hand-specific LoRA adapters
to refine finger pose. The temporal refiner produces the motion sequence;
\cref{fig:temporal_architecture} details its filtering mechanism.}
\label{fig:architecture}
\end{figure*}

\subsection{Per-Frame Whole-Body Estimator}
\label{sec:perframe}

CADENCE uses C-RADIOv4-SO400M~\cite{ranzinger2026cradiov4techreport} as a shared
visual encoder. The pretrained encoder weights are frozen. Hand-specific
LoRA~\cite{hu2022lora} adapters are enabled for hand-crop encoding. We extract early, middle,
and late visual features. Early features preserve local detail, middle features
capture limb structure, and late features provide semantic context for global
pose, shape, and camera estimation. Each feature map is projected to decoder
width $D=512$.

The decoder uses 56 parameter queries and 55 keypoint tokens, for 111 tokens
per layer. Parameter queries cover camera, shape, orientation, 21 body joints,
30 finger joints, jaw, and expression. Keypoint tokens represent where the articulated body lies in the
image. The tokens are updated jointly by a 6-layer transformer decoder with
self-attention, cross-attention to late visual features, and feed-forward blocks.
After each layer, lightweight heads predict SMPL-X updates. Pose parameters are
updated residually for coarse-to-fine refinement, while shape, expression, and
camera are predicted directly.

Let $\mathbf{Q}^{\ell}$ and $\mathbf{K}^{\ell}$ denote parameter-query and
keypoint-token states at layer $\ell$. Current parameters
$\mathbf{p}^{\ell}$ define root-relative joints and image locations,
\begin{equation}
\mathbf{J}^{\ell}=\mathrm{FK}(\mathbf{p}^{\ell}),\qquad
\mathbf{u}^{\ell}=\pi_{\mathbf{c}^{\ell}}(\mathbf{J}^{\ell}).
\end{equation}
The keypoint updater combines these coordinates with visual features sampled
at the projected locations,
\begin{equation}
\mathbf{K}^{\ell+1}=\mathcal{U}\big(\mathbf{K}^{\ell},\mathbf{J}^{\ell},
\mathbf{u}^{\ell},\mathbf{F}(\mathbf{u}^{\ell})\big).
\end{equation}
Projected joints additionally define spatial attention biases for the next
layer. From the fourth layer, wrist-centred regions supply local features to
hand queries and hand keypoint tokens. Bounding-box ray and scale information
conditions the camera and global-orientation queries. A learned embedding of
the current parameter prediction is added to the parameter queries before the
next decoder layer. This prediction-embedding path and the coordinates used to
construct attention biases are detached from gradient propagation. The spatial
feedback therefore updates the visual evidence available to later predictions.

\subsection{High-Resolution Hand Pathway}
\label{sec:hand}

Hands occupy only a small part of a full-body crop, so finger articulation is
difficult to recover from whole-body features alone. Crop-based pipelines improve
hand detail using dedicated hand encoders, but this increases complexity and can
separate the hand prediction from the global body context. CADENCE instead keeps
a shared-encoder design.

Given the current whole-body prediction, we crop the left and right hand regions
around the predicted wrists from the original image and encode them with the
same frozen C-RADIOv4 weights, adapted by a hand-specific LoRA
(rank 8, scaling 16), on $256\times256$ crops. A whole-body 2D
keypoint detector, ViTPose~\cite{xu2022vitpose}, provides the hand keypoints
$\mathbf{k}_{\mathrm{h}}$ in each crop. Crop features
$\mathbf{F}^{\mathrm{hi}}_{\mathrm{h}}$ are bilinearly sampled at these keypoints,
and a learned hand query $\mathbf{q}_{\mathrm{h}}$ cross-attends to the sampled
features to predict the finger pose of that hand:
\begin{equation}
\hat{\boldsymbol{\theta}}^{\mathrm{h}}_t =
\mathcal{H}\big(\mathbf{q}_{\mathrm{h}},
\mathbf{F}^{\mathrm{hi}}_{\mathrm{h}}(\mathbf{k}_{\mathrm{h}})\big),
\qquad
\mathrm{h}\in\{\mathrm{lh},\mathrm{rh}\}.
\end{equation}
The prediction replaces only the 15 finger rotations of each hand in
$\mathbf{p}_t$; wrist, body-pose, global-orientation, shape, expression, jaw,
and camera parameters are kept from the whole-body estimate, so body metrics are
unchanged whether the pathway runs or not. Both hands share encoder and branch
weights, and horizontal-flip augmentation covers both orientations.

Whole-body datasets often provide limited reliable supervision for fine-grained
finger articulation, so the pathway is trained as a hand-pose estimator on hand crops.
Real 3D MANO annotations come from FreiHand~\cite{freihand2019},
HanCo~\cite{hanco2021}, and ReInterHand~\cite{moon2023reinterhand}. On COCO and
UBody hand crops, where no reliable 3D ground truth exists, we distill finger
pose from HaMeR~\cite{pavlakos2024hamer}, and down-weight UBody pseudo ground truth. Keypoints are jittered during
training so that the pathway tolerates detector noise at inference.

\subsection{Kinematic-Aware Multi-Rate Temporal Refiner}
\label{sec:temporal}

\begin{figure*}[t]
\centering
\includegraphics[width=\linewidth]{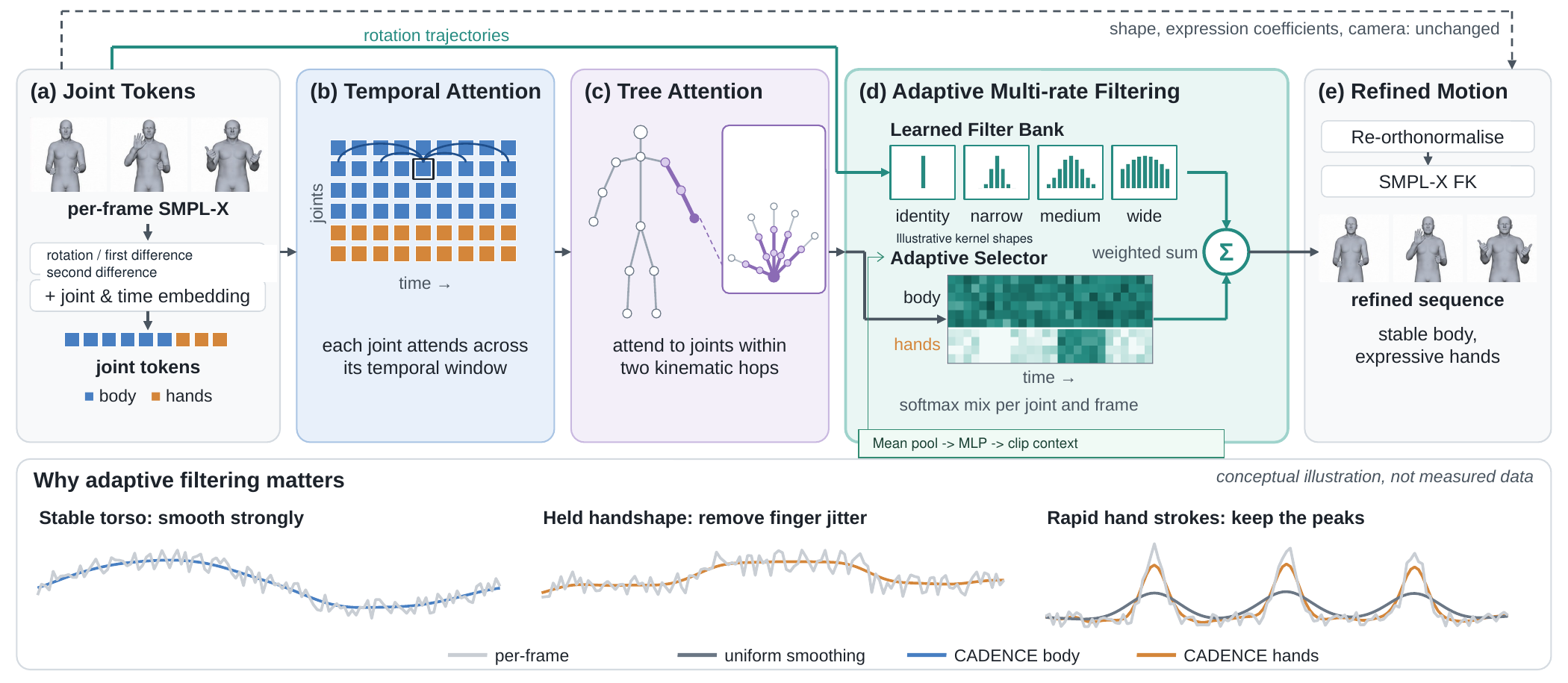}
\caption{Kinematic-aware temporal refinement. Rotation and finite-difference
features are embedded into joint tokens, processed by temporal attention and
attention within two kinematic hops, and used to select a mixture of four
learned symmetric filters for each joint and frame. Filtered rotations are
re-orthonormalised before SMPL-X forward kinematics. The lower plots illustrate
expected filtering behaviour conceptually. The selector also receives the
mean-pooled clip context of Eq.~\ref{eq:clip_context}. Its heatmap is a
conceptual summary of smoothing strength, while the model predicts four
coefficients per joint and frame.}
\label{fig:temporal_architecture}
\end{figure*}

Expressive whole-body motion requires temporal regularisation that adapts to
anatomical region and local motion. CADENCE predicts a filter mixture for each
joint and frame, conditioned on temporal and kinematic context.
Here, multi-rate denotes different effective smoothing bandwidths at the input
frame rate. Figure~\ref{fig:temporal_architecture} details the refiner.

\paragraph{Input.} The refiner takes the SMPL-X rotation sequence for a clip of
$T$ frames and reads out $J{=}53$ per-frame rotations: global orientation, the
21 body joints, the jaw, and both hands. Each rotation $\mathbf{r}_{j,t}\in\mathbb{R}^6$
is embedded together with its first-order temporal difference
$\mathbf{r}_{j,t}-\mathbf{r}_{j,t-1}$ and second-order temporal difference
$\mathbf{r}_{j,t+1}-2\mathbf{r}_{j,t}+\mathbf{r}_{j,t-1}$, a learned joint
identity, and a temporal position embedding, giving a token
$\mathbf{h}_{j,t}\in\mathbb{R}^d$. The refiner uses pose only. Its embedding width is $d=192$, with two temporal
transformer layers followed by two tree-attention layers, each with four heads
and feed-forward width $4d$.

\paragraph{Kinematic-tree attention.} Tokens are first updated by self-attention
over time, one joint at a time, and then by self-attention over joints within a
frame, masked to the SMPL-X kinematic tree:
\begin{equation}
\begin{gathered}
\mathbf{h}_{:,t} \leftarrow \mathrm{Attn}(\mathbf{h}_{:,t}, \mathbf{h}_{:,t}, \mathbf{h}_{:,t}; M),\\
M_{jk} = \begin{cases} 0 & d_{\mathrm{tree}}(j,k) \le 2, \\ -\infty & \text{otherwise,} \end{cases}
\end{gathered}
\end{equation}
where $d_{\mathrm{tree}}$ is the number of edges between joints $j$ and $k$ in
the tree. A wrist token thus sees its elbow and its finger joints in one pass, so
the bandwidth chosen for a finger can depend on whether the arm is moving.

\paragraph{Content-adaptive multi-rate filtering.} A bank of $F{=}4$ learned
kernels $\mathbf{w}_f\in\mathbb{R}^{K}$ ($K{=}9$) is applied to every joint's
rotation track. Each kernel is symmetric, which centres its temporal support, and is
softmax-normalised, which gives non-negative weights summing to one:
\begin{equation}
\tilde{\mathbf{r}}^{(f)}_{j,t} = \sum_{k=-4}^{4} w_{f,k}\, \mathbf{r}_{j,t+k}.
\end{equation}
The kernels are initialised to the identity and to Gaussians with
$\sigma{=}1, 2, 4$ frames. The constraints do not enforce an ordering of
learned bandwidths or a monotonic frequency response. The learned kernels and
their responses are shown in \cref{fig:learned_filters}. For every joint and frame, a selector predicts mixing
weights over the bank from the token and a clip-level motion context
$\mathbf{c}$, computed by pooling over joints and frames,
\begin{equation}
\mathbf{c}=\mathrm{MLP}\left(\frac{1}{JT}\sum_{j,t}\mathbf{h}_{j,t}\right).
\label{eq:clip_context}
\end{equation}
The selector and refined rotations are then
\begin{equation}
\begin{aligned}
\boldsymbol{\alpha}_{j,t} &= \mathrm{softmax}\big(\mathbf{W}(\mathbf{h}_{j,t}+\mathbf{c}) + \mathbf{b}_j\big),\\
\hat{\mathbf{r}}_{j,t} &= \sum_f \alpha_{j,t,f}\,\tilde{\mathbf{r}}^{(f)}_{j,t},
\end{aligned}
\end{equation}
followed by Gram-Schmidt re-orthonormalisation. This per-joint, per-frame choice
of bandwidth is what we mean by \emph{multi-rate}. The per-joint bias
$\mathbf{b}_j$ is initialised to select the $\sigma{=}2$ kernel for body and
orientation joints and the identity kernel for the jaw, forearms, and hands, and
$\mathbf{W}$ is zero-initialised. This establishes anatomically differentiated
temporal filtering before optimisation. The effective local weights remain symmetric, non-negative, and normalised.
These constraints provide centred averaging in 6D rotation space. Adaptive
selection and rotation projection can still alter articulated trajectories. When a hand holds a handshape, the selector can move to a smoother
kernel; during a stroke it can stay near the identity. The single-rate variant
in \cref{sec:temporal_results} shares one kernel across all joints.

The refiner operates offline on 32-frame windows with a hop of 16 frames,
blended with a Hann window; shape, expression, and camera pass through unchanged.

\subsection{Training Objectives}
\label{sec:training}

The per-frame estimator is supervised at each decoder layer with a combination of
rotation, 3D joint, 2D keypoint, vertex, camera, shape, and expression losses:
\begin{equation}
\begin{aligned}
\mathcal{L}_{\mathrm{pf}} ={}&
\lambda_{\mathrm{rot}}\mathcal{L}_{\mathrm{rot}}
+\lambda_{\mathrm{j3d}}\mathcal{L}_{\mathrm{j3d}}
+\lambda_{\mathrm{j2d}}\mathcal{L}_{\mathrm{j2d}}
+\lambda_{\mathrm{v}}\mathcal{L}_{\mathrm{v}} \\
&+\lambda_{\mathrm{cam}}\mathcal{L}_{\mathrm{cam}}
+\lambda_{\beta}\mathcal{L}_{\beta}
+\lambda_{\psi}\mathcal{L}_{\psi}.
\end{aligned}
\end{equation}
The hand pathway is supervised with geodesic finger-rotation losses.
FreiHand, HanCo, and ReInterHand provide 3D rotations. COCO and UBody crops
use HaMeR rotation distillation, with an additional down-weighted pseudo-pose
anchor on UBody. Global hand orientation is supervised where reliable during
hand-only training, while inference updates the finger rotations.

The temporal refiner is trained from stored predictions with the image-based
components frozen. Reconstruction terms are combined with output-acceleration
regularisation and velocity-preservation penalties on high-speed frames.
Body and hand acceleration terms have weights 8 and 2. Peak-preservation
penalties protect the highest-speed frames, using quantiles 0.85 and 0.70 and
minimum retained-speed fractions 0.78 and 0.92 for body and hands.
Checkpoint selection constrains validation MPJPE regression to 0.2\,mm,
requires non-increasing acceleration error, and limits per-clip regression to
2\,mm. This validation criterion is the accuracy safeguard used during
refiner training.

Training fits the image-based components before temporal refinement.
The hand adapters are active only on hand crops, preserving the whole-body
encoder configuration used to produce the stored temporal inputs.
\Cref{sec:supp_impl_training} gives implementation details.

\section{Experiments}
\label{sec:experiments}

\subsection{Experimental Setup}
\label{sec:setup}

\paragraph{Datasets.}
The per-frame estimator is trained on AGORA~\cite{Patel:CVPR:2021},
BEDLAM~\cite{Black_CVPR_2023}, UBody~\cite{lin2023one},
Motion-X~\cite{zhang2025motion}, and 3DPW~\cite{vonmarcard2018}, with 2D
keypoint supervision from COCO-WholeBody~\cite{jin2020whole} and
MPII~\cite{andriluka14cvpr}. We add 500K frames of public 2D images from
COCO-WholeBody, MPII, and InstaVariety~\cite{kanazawa2019learning} with
pseudo-SMPL-X labels from the public SMPLer-X-L~\cite{cai2023smplerx}
checkpoint, rejecting frames whose SMPLer-X 2D reprojection error exceeds 20
pixels; no test images are used. The hand pathway uses the hand data in
\cref{sec:hand}, and the temporal refiner trains on clips from the video
sources. \Cref{sec:supp_training_data}
lists per-source counts.

\paragraph{Metrics.}
MPJPE and MPVPE (also written PVE or MVE) are mean Euclidean joint and vertex
errors in millimetres; the PA- and P- prefixes denote errors after a
similarity Procrustes alignment. For EHF~\cite{SMPL-X:2019}, hand and face
mesh errors are computed after wrist and neck alignment, following
SMPLer-X~\cite{cai2023smplerx}. 3DPW uses the standard 14-joint protocol and
EMDB the 24-joint protocol; neither measures hands, absolute depth, or global
placement. Acceleration error (Accel) is the mean difference between
predicted and ground-truth joint accelerations, computed by central second
differences over valid consecutive frames, in mm/frame$^2$, without multiplication by the squared frame rate. On sign-language
sequences, Jitter is the mean norm of the predicted joints' second difference,
$\operatorname{mean}_{t,j}\lVert\mathbf{p}_{t+2,j}-2\mathbf{p}_{t+1,j}+\mathbf{p}_{t,j}\rVert_2$,
in mm/frame$^2$ over valid three-frame runs, with body joints
relative to the pelvis and hand joints relative to the wrist. It is a prediction-only smoothness proxy that also responds to genuine
articulation. Peak-R is the fraction
of reference hand-velocity peaks that the prediction recovers, and Peak-A is the
ratio of predicted to reference peak hand-velocity amplitude. A value of 1 means matching
peak size, values below 1 indicate attenuated strokes, and values above 1
overshoot. The sign-language reference motion is pseudo ground truth and contains jitter
of its own, so we use these metrics to compare smoothers with each other rather
than as absolute accuracy.

\paragraph{Implementation.}
The C-RADIOv4-SO400M encoder~\cite{ranzinger2026cradiov4techreport} has about
472M parameters and is frozen except for LoRA adapters used during hand-pathway training. The per-frame estimator has 51.8M trainable parameters, the
hand pathway 21.0M, and the temporal refiner 1.88M. Person crops are
$384\times288$. One model is used for all benchmarks, without per-benchmark
fine-tuning. We chose C-RADIOv4 because it is distilled from several
foundation models, including DINOv3 and SAM, which lets the whole-body decoder
and the hand pathway share one backbone; we have not compared it against a
human-specific backbone.

\subsection{Temporal Coherence}
\label{sec:temporal_results}

We evaluate whether the refiner stabilises whole-body motion without
erasing fast hand motion. All smoothers below take the same stored per-frame
CADENCE predictions as input, so the rows differ only in temporal processing.
Single-rate is the CADENCE refiner with one temporal bandwidth shared by all
joints. SmoothNet-WB is SmoothNet~\cite{zeng2022smoothnet} retrained on the full
341-dimensional SMPL-X vector, including the hand channels.

\begin{table}[!t]
\centering
\small
\caption{Temporal smoothing on 3DPW. Every method smooths the same per-frame
CADENCE predictions. Accel in mm/frame$^2$, errors in mm.}
\label{tab:temporal_smoothing}
\setlength{\tabcolsep}{5pt}
\resizebox{\linewidth}{!}{%
\begin{tabular}{@{}lccc@{}}
\toprule
Method & PA-MPJPE $\downarrow$ & MPJPE $\downarrow$ & Accel $\downarrow$ \\
\midrule
Per-frame (no smoothing) & 41.9 & 61.7 & 17.4 \\
\midrule
Gaussian ($\sigma{=}2$) & 41.5 & 61.1 & 7.0 \\
Moving average ($w{=}5$) & 41.5 & 61.1 & 6.6 \\
One-Euro~\cite{10.1145/2207676.2208639} & 50.5 & 70.3 & 6.2 \\
Savitzky--Golay (order 2, $w{=}7$) & 41.6 & 61.4 & 7.6 \\
SmoothNet-WB~\cite{zeng2022smoothnet} & 42.5 & 61.9 & 7.7 \\
Single-rate refiner & 41.7 & 61.2 & 5.7 \\
\rowcolor{green!10}
CADENCE refiner & \textbf{40.2} & \textbf{58.1} & \textbf{5.4} \\
\bottomrule
\end{tabular}%
}

\end{table}

\begin{table}[!t]
\centering
\small
\caption{Temporal stability and hand-motion preservation on
SignAvatars/How2Sign~\cite{yu2024signavatars} test sequences. Joint errors are
in mm and jitter in mm/frame$^2$. Peak-R measures reference peak recall and
Peak-A the reference-relative peak velocity amplitude ratio (ideal 1).
Reference motion is obtained from pseudo-ground-truth annotations.}
\label{tab:sign_dynamics}
\setlength{\tabcolsep}{3pt}
\resizebox{\linewidth}{!}{%
\begin{tabular}{@{}lcccccc@{}}
\toprule
Method & PA-MPJPE $\downarrow$ & MPJPE $\downarrow$ & Jitter $\downarrow$
& H-MPJPE $\downarrow$ & Peak-R $\uparrow$ & Peak-A (ideal 1) \\
\midrule
Per-frame & 10.72 & 12.64 & 8.021 & 19.04 & 0.93 & 1.18 \\
\midrule
SmoothNet-WB & 10.13 & 11.95 & 1.590 & 17.82 & 0.81 & 0.35 \\
Gaussian ($\sigma{=}2$) & 10.32 & 12.20 & 1.426 & 18.16 & 0.62 & 0.49 \\
One-Euro & \textbf{9.17} & \textbf{10.78} & 1.894 & \textbf{17.24} & 0.83 & 0.66 \\
Moving avg.\ ($w{=}5$) & 10.46 & 12.34 & 2.221 & 18.41 & 0.78 & 0.58 \\
Savitzky--Golay & 10.66 & 12.56 & 3.250 & 18.82 & 0.83 & 0.76 \\
Single-rate refiner & 10.35 & 12.23 & 1.497 & 18.21 & 0.62 & 0.50 \\
\rowcolor{green!10}
CADENCE refiner & 10.28 & 12.16 & \textbf{1.286} & 18.68 & \textbf{0.90} & \textbf{0.98} \\
\bottomrule
\end{tabular}%
}
\end{table}

On 3DPW (\cref{tab:temporal_smoothing}), the refiner reduces acceleration
error from 17.4 to 5.4\,mm/frame$^2$, the lowest in the table, and improves pose
accuracy more than any other row: PA-MPJPE falls from 41.9 to 40.2\,mm and
MPJPE from 61.7 to 58.1\,mm. The zero-phase classical filters reach 6.6 to
7.6\,mm/frame$^2$ with small accuracy changes. The One-Euro configuration reaches
6.2\,mm/frame$^2$, with PA-MPJPE increasing to 50.5\,mm. SmoothNet-WB
reduces acceleration error to 7.7\,mm/frame$^2$ while slightly increasing pose
error.

\begin{figure*}[!t]
\centering
\includegraphics[width=0.76\linewidth]{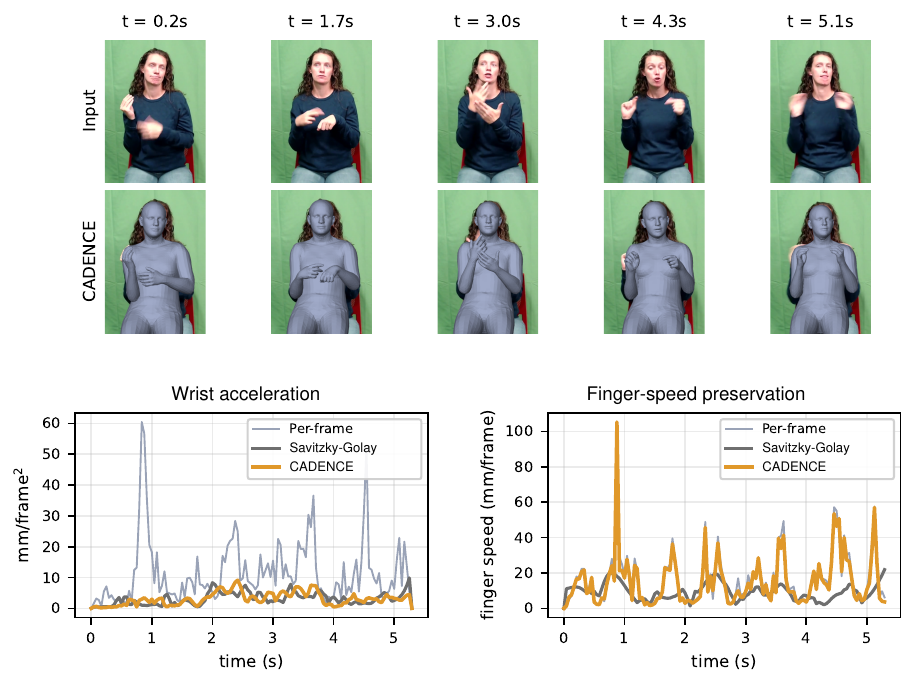}
\caption{Matched-jitter comparison on one sign-language clip. Savitzky--Golay
and CADENCE reach wrist jitter of 2.5 and 2.6\,mm/frame$^2$, respectively.
CADENCE retains 99\% of the per-frame peak finger speed and the matched
smoother retains 46\%. These single-clip ratios use the per-frame prediction
as reference, while \cref{tab:sign_dynamics} uses pseudo ground truth.}
\label{fig:supp_temporal_qual}
\end{figure*}

3DPW measures body joints only, so it cannot show what smoothing does to the
hands. We therefore evaluate hand dynamics in \cref{tab:sign_dynamics}. Every smoother reduces hand jitter
relative to the per-frame input (8.02), and the refiner reaches the lowest value
(1.29). The methods exhibit different stability--preservation trade-offs. SmoothNet-WB
and the Gaussian filter keep only 0.35 and 0.49 of the peak hand speed; the
refiner keeps 0.98 and recovers 0.90 of the peaks, against 0.93 for the
unsmoothed input. The single-rate refiner reaches jitter similar to the
Gaussian filter (1.50) and keeps a similar peak amplitude (0.50), so the
difference between the two refiner rows comes from the per-joint bandwidth.
The One-Euro filter has the lowest pose errors on this benchmark but keeps less
peak amplitude (0.66) at higher jitter (1.89). On a single sign-language clip
where a Savitzky--Golay filter is tuned to match the refiner's residual wrist
jitter, the refiner keeps 99\% of the per-frame peak finger speed and the
matched filter 46\% (\cref{fig:supp_temporal_qual}). Among the compared
smoothers, the refiner therefore gives the lowest hand jitter while keeping
nearly all hand-speed peak amplitude.

Fixed-filter comparisons on matched clip subsets (\cref{sec:supp_temporal})
further demonstrate that joint-dependent filtering improves hand-velocity
peak preservation at comparable temporal smoothness.

\begin{figure}[!t]
\centering
\includegraphics[width=\linewidth]{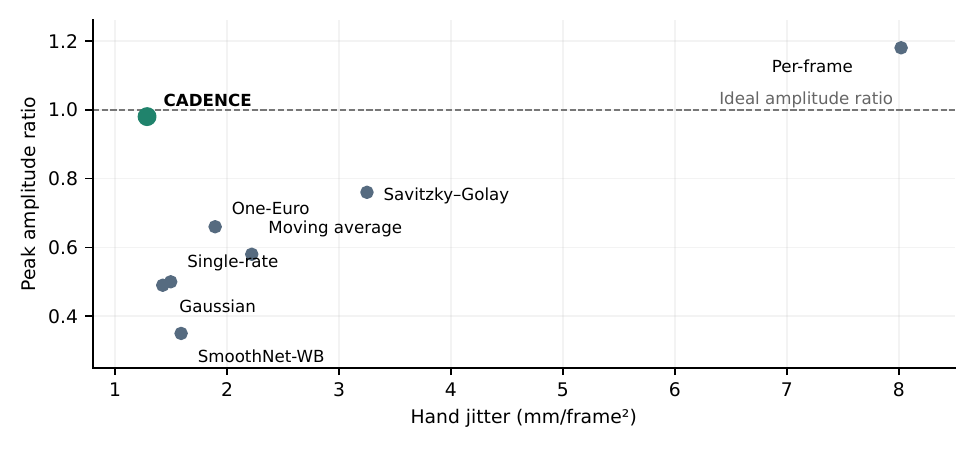}
\caption{Hand jitter and peak amplitude ratio from \cref{tab:sign_dynamics}.
Lower jitter and an amplitude ratio near one are desirable. Points represent
individual evaluated configurations, without a smoothing-strength sweep.}
\label{fig:stability_preservation}
\end{figure}

\subsection{Body Accuracy and Smoothness on 3DPW and EMDB}
\label{sec:body_results}

\Cref{tab:body_compact} compares CADENCE with published methods on the
body-only protocols of 3DPW and EMDB. Acceleration values are reported for
CADENCE using the local unscaled second-difference protocol. Dashes mark
unreported or protocol-unverified measurements. The matched-input experiment in
\cref{tab:temporal_smoothing} provides the controlled temporal comparison.

\begin{table*}[!t]
\centering
\small
\caption{In-the-wild body benchmarks (3DPW 14-joint protocol, EMDB 24-joint
protocol). Errors in mm, Accel in mm/frame$^2$. CADENCE predicts whole-body SMPL-X;
some reference methods are body-only. Results of other methods are as reported
in their papers; DanceHMR has no public code. Best results are \textbf{bold},
second-best \underline{underlined}.}
\label{tab:body_compact}
\setlength{\tabcolsep}{4.0pt}
\resizebox{\textwidth}{!}{%
\begin{tabular}{@{}llccccccc@{}}
\toprule
& & \multicolumn{4}{c}{3DPW} & \multicolumn{3}{c}{EMDB} \\
\cmidrule(lr){3-6}\cmidrule(lr){7-9}
Method & Repr.
& PA-MPJPE $\downarrow$ & MPJPE $\downarrow$ & PVE $\downarrow$ & Accel $\downarrow$
& PA-MPJPE $\downarrow$ & MPJPE $\downarrow$ & PVE $\downarrow$ \\
\midrule
\multicolumn{9}{@{}l}{\textit{Whole-body SMPL-X methods}} \\
\quad Multi-HMR~\cite{multi-hmr2024} & SMPL-X & 41.7 & 61.4 & -- & -- & -- & -- & -- \\
\quad DanceHMR~\cite{shen2026dancehmr} & SMPL-X video
& 35.2 & \textbf{53.6} & 65.1 & -- & 40.1 & \underline{64.3} & \underline{75.4} \\
\quad CADENCE per-frame & SMPL-X & 41.9 & 61.7 & -- & 17.4 & -- & -- & -- \\
\rowcolor{green!10}
\quad CADENCE & SMPL-X video
& 40.2 & 58.1 & \textbf{62.6} & 5.4 & \textbf{37.2} & 69.5 & 76.0 \\
\midrule
\multicolumn{9}{@{}l}{\textit{Body-focused per-frame or fitting methods}} \\
\quad PromptHMR~\cite{wang2025prompthmr} & SMPL-X & 36.1 & 58.7 & 69.4 & -- & 41.0 & 71.7 & 84.5 \\
\quad NLF-L+fit~\cite{sarandi2024nlf} & SMPL & \underline{33.6} & 54.9 & 63.7 & -- & 40.9 & 68.4 & 80.6 \\
\quad SAM 3D Body~\cite{sam3dbody} & Body & \textbf{33.2} & 54.8 & \underline{63.6} & -- & \underline{38.2} & \textbf{61.7} & \textbf{72.5} \\
\midrule
\multicolumn{9}{@{}l}{\textit{Body-only temporal methods}} \\
\quad TCMR~\cite{choi2020beyond} & SMPL video & 55.8 & 95.0 & -- & -- & -- & -- & -- \\
\quad MPS-Net~\cite{WeiLin2022mpsnet} & SMPL video & 52.1 & 84.3 & -- & -- & -- & -- & -- \\
\quad WHAM~\cite{wham:cvpr:2024} & SMPL video & 35.9 & 57.8 & 68.7 & -- & -- & -- & -- \\
\quad TRAM~\cite{wang2024tram} & SMPL video & 35.6 & 59.3 & 69.6 & -- & -- & -- & -- \\
\quad GVHMR~\cite{shen2024gvhmr} & SMPL video & 36.2 & 55.6 & 67.2 & -- & -- & -- & -- \\
\quad GENMO~\cite{li2025genmo} & SMPL video & 34.6 & \underline{53.9} & 65.8 & -- & 42.5 & 73.0 & 84.8 \\
\bottomrule
\end{tabular}%
}
\end{table*}

CADENCE has the lowest 3DPW PVE (62.6\,mm). Its joint errors are higher than those of the
strongest body-focused methods: on 3DPW, SAM~3D~Body and NLF-L+fit reach 33.2
and 33.6\,mm PA-MPJPE against CADENCE's 40.2\,mm, and DanceHMR reaches
53.6\,mm MPJPE against 58.1\,mm. On EMDB, CADENCE has the lowest PA-MPJPE
(37.2\,mm) but higher MPJPE and PVE than SAM~3D~Body and DanceHMR, suggesting that global alignment remains challenging relative to aligned
articulated structure. These protocols score body joints only and do not measure the hand
and face output.

\subsection{Expressive Whole-Body Accuracy on EHF}
\label{sec:ehf}

\Cref{tab:ehf_compact} reports EHF, the standard expressive whole-body SMPL-X
benchmark. CADENCE has the lowest full-body P-MVE (25.7\,mm), body P-MPJPE
(20.6\,mm), and hand P-MPJPE (8.3\,mm) among the listed methods. Its hand mesh
error (12.2\,mm) is higher than that of PyMAF-X (10.2\,mm), which uses dedicated
hand crops and a separate hand branch, and Multi-HMR (10.8\,mm); SMPLer-X-L has
the lowest face P-MVE. Lower joint error with higher mesh error suggests that the
pathway recovers finger articulation better than finger surface shape.

\begin{table}[!t]
\centering
\small
\caption{EHF~\cite{SMPL-X:2019} whole-body evaluation (mm). Hand and face errors
are wrist- and neck-aligned. The whole-body and hand-refined configurations are reported separately.
Dashes indicate configuration-specific measurements that are unavailable. Best \textbf{bold}, second \underline{underlined}.
\Cref{sec:supp_hand_ablation} gives hand pathway diagnostics.}
\label{tab:ehf_compact}
\setlength{\tabcolsep}{4pt}
\resizebox{\linewidth}{!}{%
\begin{tabular}{@{}lccccc@{}}
\toprule
& \multicolumn{3}{c}{P-MVE $\downarrow$} & \multicolumn{2}{c}{P-MPJPE $\downarrow$} \\
\cmidrule(lr){2-4}\cmidrule(lr){5-6}
Method & Full & Hand & Face & Body & Hand \\
\midrule
PyMAF-X~\cite{pymafx2023} & 50.2 & \textbf{10.2} & 5.5 & 52.8 & \underline{10.3} \\
SMPLer-X-L~\cite{cai2023smplerx} & 37.1 & 14.1 & \textbf{5.0} & -- & -- \\
Multi-HMR~\cite{multi-hmr2024} & 28.2 & \underline{10.8} & 5.3 & -- & -- \\
NLF-L+fit~\cite{sarandi2024nlf} & \underline{26.0} & 20.7 & 6.3 & \underline{26.1} & 21.8 \\
\midrule
\rowcolor{green!10}
CADENCE whole-body & \textbf{25.7} & 22.6 & \underline{5.1} & \textbf{20.6} & 14.5 \\
CADENCE + hands & -- & 12.2 & -- & -- & \textbf{8.3} \\
\bottomrule
\end{tabular}%
}

\end{table}

\begin{figure*}[!t]
\centering
\setlength{\tabcolsep}{1pt}
\renewcommand{\arraystretch}{0.5}
\begin{tabular}{cccc}
\includegraphics[width=0.245\textwidth]{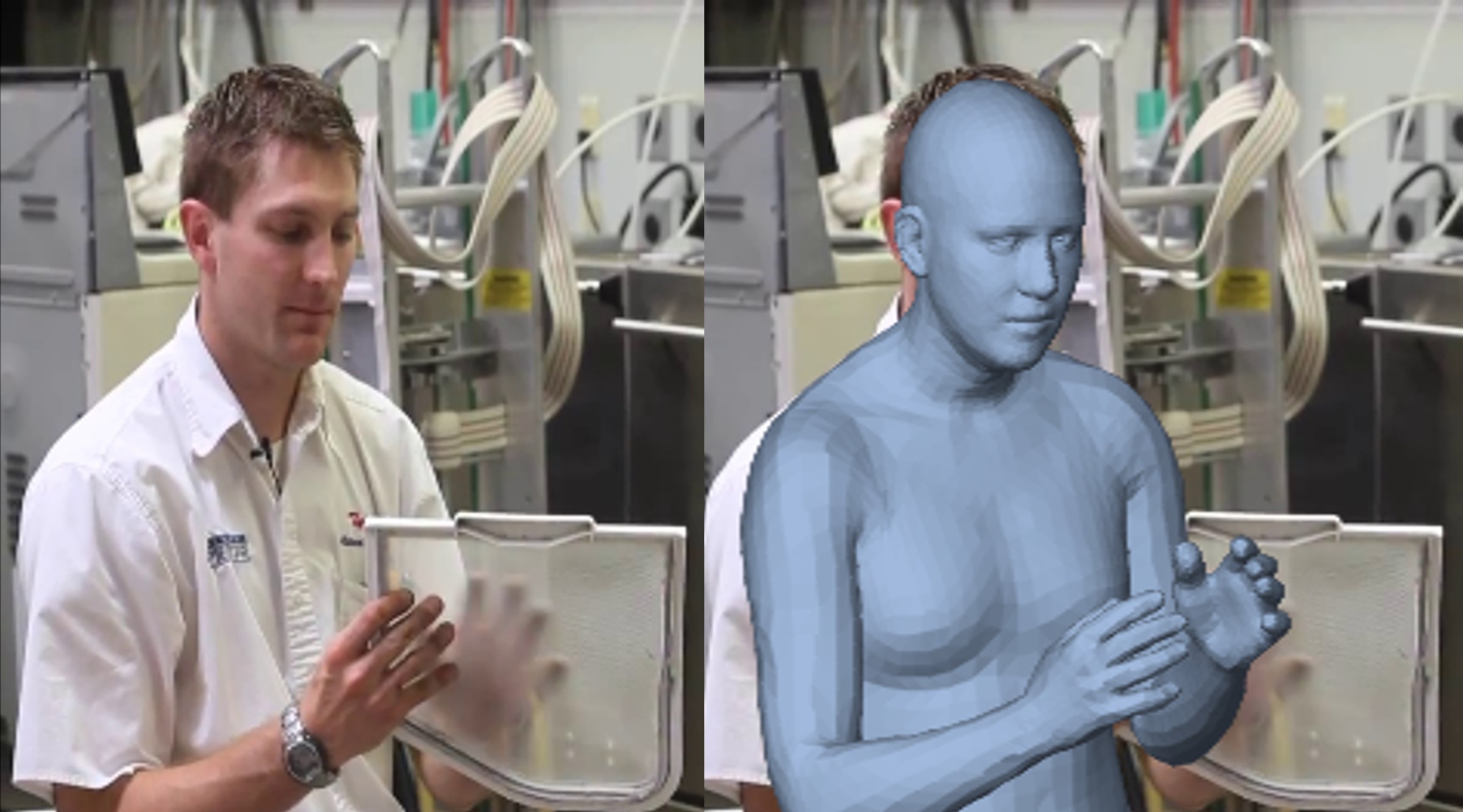} &
\includegraphics[width=0.245\textwidth]{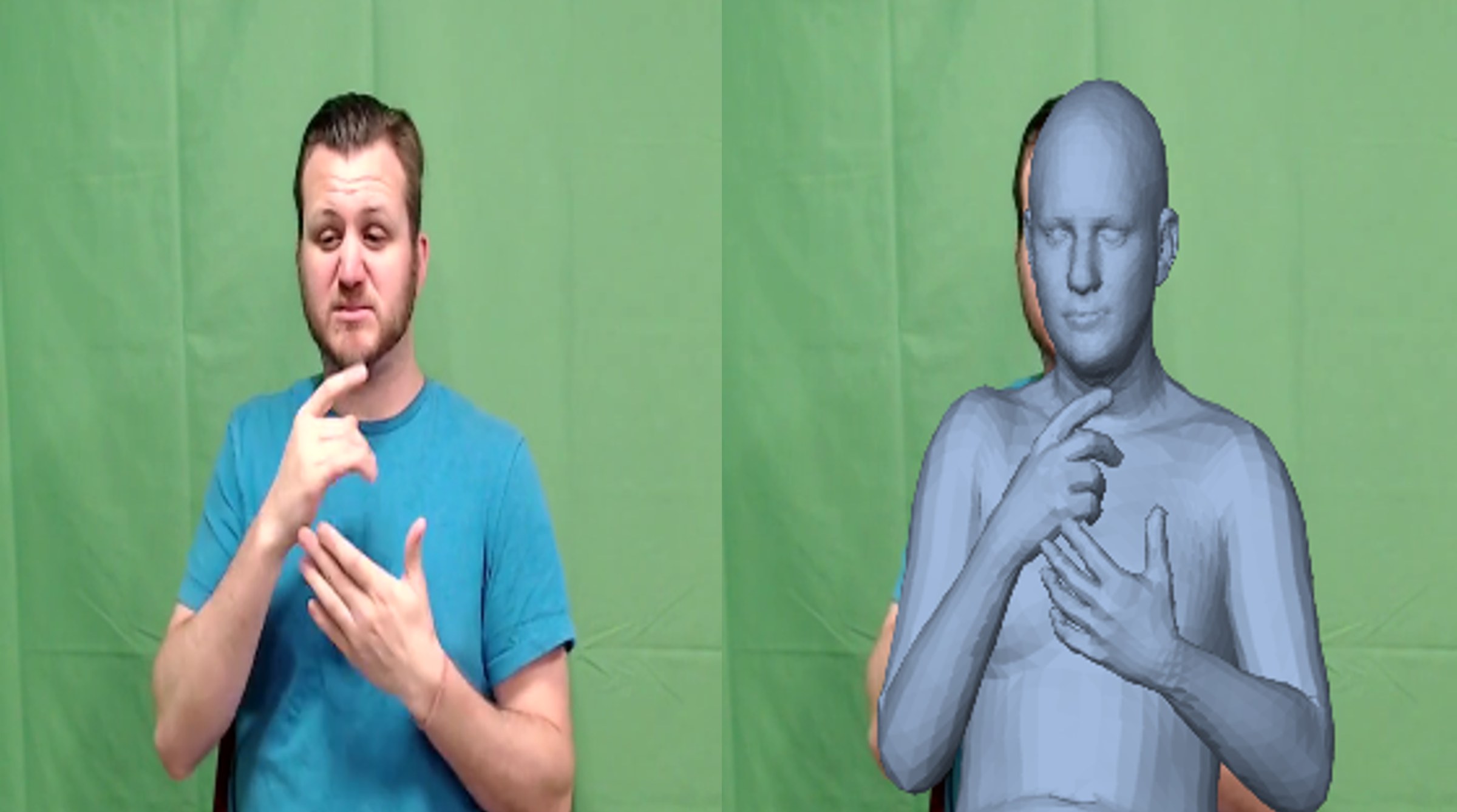} &
\includegraphics[width=0.245\textwidth]{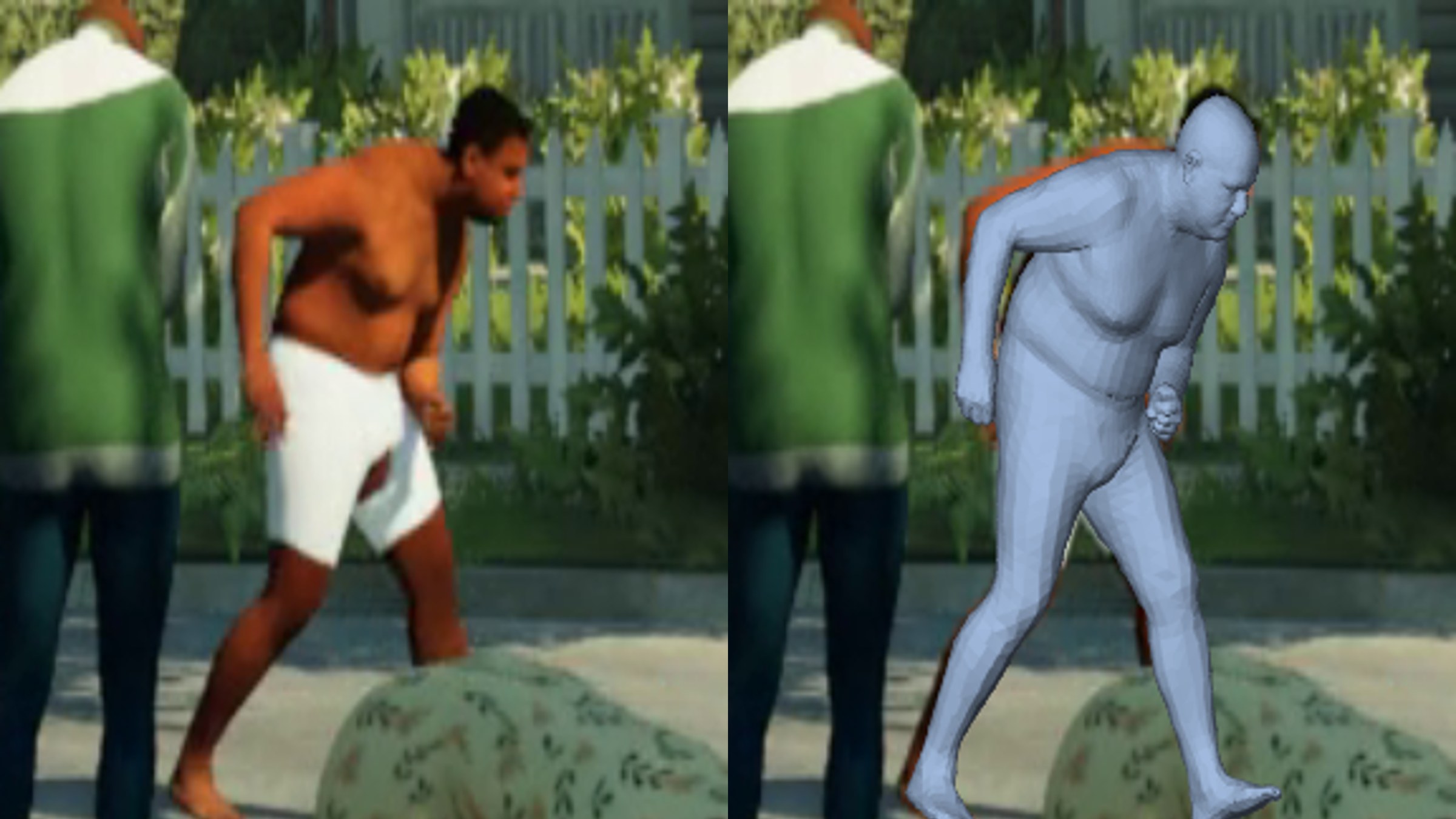} &
\includegraphics[width=0.245\textwidth]{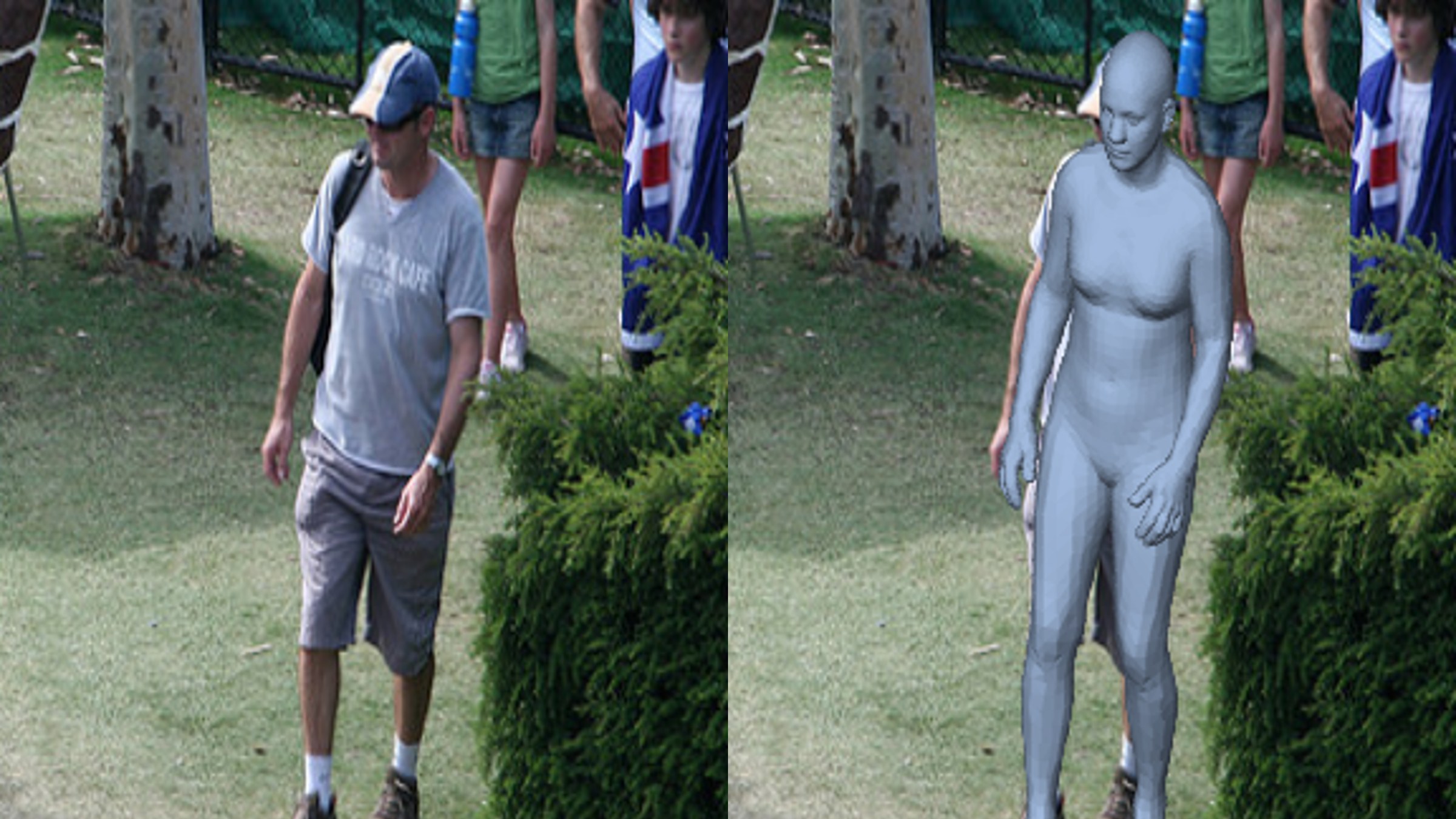} \\
\end{tabular}
\caption{CADENCE reconstructions on samples from 3DPW~\cite{vonmarcard2018},
UBody~\cite{lin2023one}, BEDLAM~\cite{Black_CVPR_2023}, and in-the-wild images,
covering indoor and outdoor scenes, seated and standing poses, sports motion,
occlusion, and expressive upper-body motion. \Cref{sec:supp_qual_failure}
shows further qualitative comparisons.}
\label{fig:qualitative_main}
\end{figure*}

\subsection{Whole-Body Generalisation on AGORA and UBody}
\label{sec:generalization}

\Cref{tab:wholebody_generalization} evaluates whole-body mesh error on AGORA
validation and UBody. CADENCE has the lowest full-body PA-PVE and PVE on both
datasets and the lowest UBody hand PA-PVE (5.9\,mm). DanceHMR has the lowest
UBody hand PVE (22.2 against 28.2\,mm); since that metric is not
Procrustes-aligned, it also reflects wrist placement and complements the
hand-aligned reconstruction metric.

\begin{table}[!t]
\centering
\small
\caption{Whole-body mesh error (mm) on AGORA~\cite{Patel:CVPR:2021} validation
and UBody~\cite{lin2023one}. CADENCE uses the whole-body estimator alone,
evaluated frame by frame on both datasets. Best per column \textbf{bold}, second
\underline{underlined}. DanceHMR values are as reported.}
\label{tab:wholebody_generalization}
\setlength{\tabcolsep}{2.5pt}
\resizebox{\linewidth}{!}{%
\begin{tabular}{@{}lcccccc@{}}
\toprule
& \multicolumn{2}{c}{AGORA val} & \multicolumn{4}{c}{UBody} \\
\cmidrule(lr){2-3}\cmidrule(lr){4-7}
& \multicolumn{2}{c}{All} & \multicolumn{2}{c}{All} & \multicolumn{2}{c}{Hand} \\
\cmidrule(lr){2-3}\cmidrule(lr){4-5}\cmidrule(lr){6-7}
Method & PA-PVE & PVE & PA-PVE & PVE & PA-PVE & PVE \\
\midrule
OSX~\cite{lin2023one} & 69.4 & 168.6 & 42.4 & 92.4 & 10.8 & 47.7 \\
SMPLer-X-L~\cite{cai2023smplerx} & \underline{45.1} & \underline{74.2} & 33.2 & 61.5 & 10.6 & 43.3 \\
DanceHMR~\cite{shen2026dancehmr} & -- & -- & \underline{31.3} & \underline{61.0} & \underline{7.4} & \textbf{22.2} \\
\midrule
\rowcolor{green!10}
CADENCE whole-body & \textbf{42.4} & \textbf{70.8} & \textbf{29.3} & \textbf{58.3} & \textbf{5.9} & \underline{28.2} \\
\bottomrule
\end{tabular}%
}
\end{table}

\subsection{Distribution Shift on Hand-Object Video}
\label{sec:arctic}

EHF, AGORA, and UBody use fitted or pseudo ground truth for hands.
ARCTIC~\cite{fan2023arctic} provides accurate 3D hand ground truth on video of
two-handed object manipulation, a setting absent from CADENCE's training data.
We evaluate the unchanged model on 8 held-out sequences (296 frames), using
crops built from ground-truth joint boxes because ARCTIC's own crops are too
tight for a whole-body model. CADENCE reaches body and hand PA-MPJPE of 49.80
and 32.02\,mm, well above its error on EHF, AGORA, and UBody. Close-range
manipulation and object occlusion are outside the training distribution, and
this result is consistent with the occlusion sensitivity measured on AGORA
(\cref{sec:runtime_analysis}).

\subsection{Effect of High-Resolution Hand Refinement}
\label{sec:hand_ablation}

\begin{figure}[!t]
\centering
\includegraphics[width=0.72\linewidth]{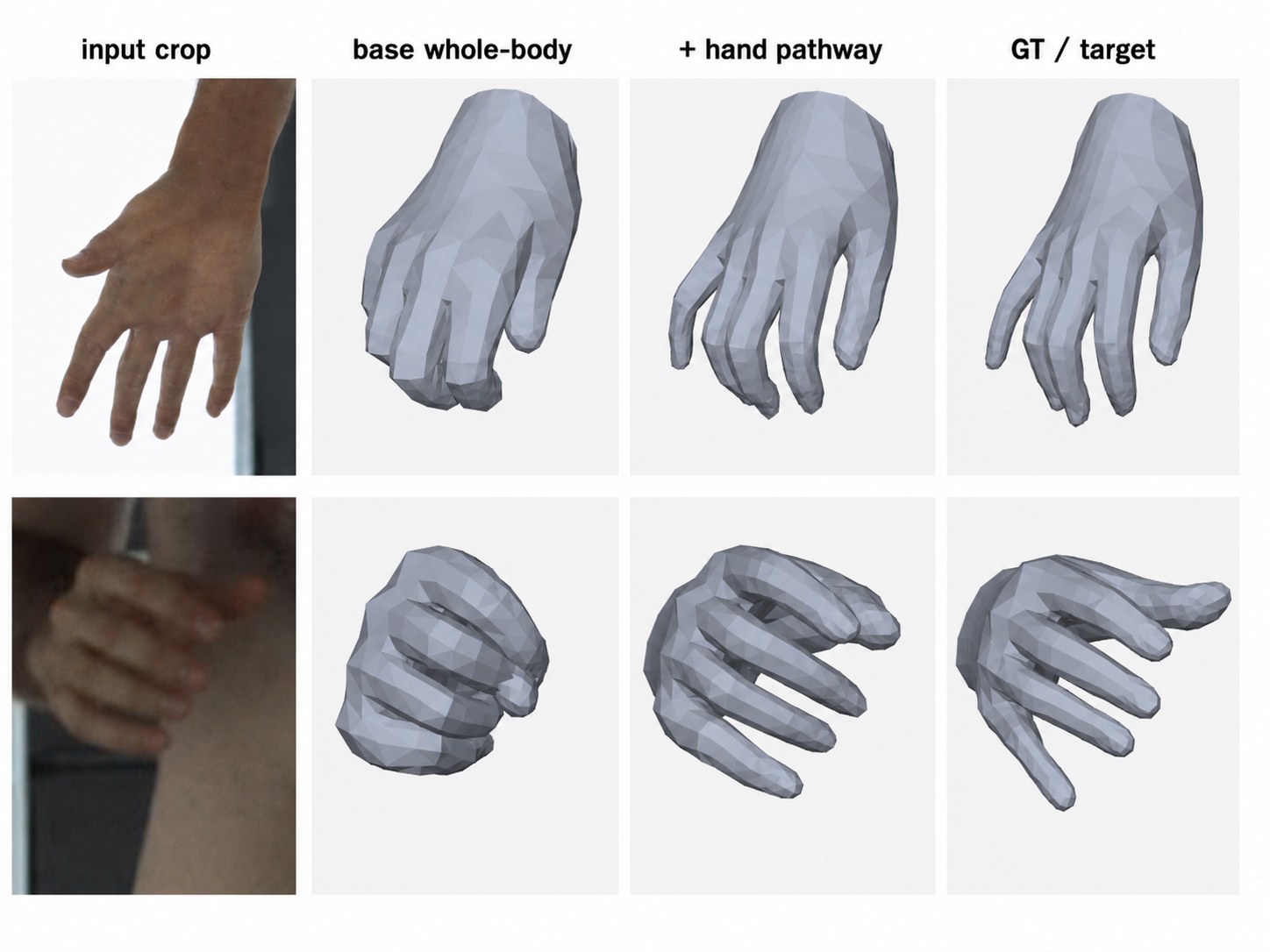}
\caption{High-resolution hand refinement. Each row presents an input hand crop, the corresponding whole-body estimate,
the hand-refined estimate, and the reference mesh.}
\label{fig:supp_hand_pathway}
\end{figure}

\Cref{tab:hand_ablation} compares hand refinement configurations on EHF
with the whole-body estimate fixed. Crop-based residual refinement reduces
hand P-MVE from 22.6 to 16.6\,mm. Adding FreiHand and HanCo supervision yields
16.5\,mm. CADENCE's keypoint-guided pathway achieves 12.2\,mm P-MVE and
8.3\,mm P-MPJPE, reductions of 46.0\% and 42.8\% relative to whole-body
estimation alone. These configurations vary architecture and supervision
jointly. The controlled-budget comparisons in \cref{sec:supp_hand_ablation}
provide further analysis, and \cref{fig:supp_hand_pathway} shows qualitative
examples.

\begin{table}[!t]
\centering
\small
\caption{Hand refinement architectures and supervision on EHF (mm).
The keypoint-guided configuration combines architectural and supervision changes.}
\label{tab:hand_ablation}
\setlength{\tabcolsep}{4pt}
\begin{tabular}{@{}lcc@{}}
\toprule
& \multicolumn{2}{c}{Hand} \\
\cmidrule(lr){2-3}
Configuration & P-MVE $\downarrow$ & P-MPJPE $\downarrow$ \\
\midrule
Whole-body estimation only & 22.6 & 14.5 \\
Residual crop pathway, pseudo-GT & 16.6 & 11.1 \\
\quad + FreiHand/HanCo & 16.5 & 11.0 \\
\rowcolor{green!10}
Keypoint-guided pathway & \textbf{12.2} & \textbf{8.3} \\
\bottomrule
\end{tabular}
\end{table}

\subsection{Runtime and Occlusion}
\label{sec:runtime_analysis}

On one A100-80GB GPU in bf16, the per-frame estimator runs at 27.9\,FPS at
batch size~1 and 307.6\,FPS at batch size~32. With the hand pathway, which
encodes two extra crops per frame, the full system runs at 13.1 and 71.5\,FPS.
The temporal refiner processes a 32-frame window at about 6{,}900\,FPS, so its
cost is negligible, and peak memory stays below 4\,GiB
(\cref{tab:supp_runtime_accounting}). Applications that do not require fine-grained finger
reconstruction can disable the hand pathway and keep the per-frame rate.

To measure sensitivity to occlusion, we split 500 AGORA validation samples by
their annotated occlusion percentage, which on AGORA comes mostly from other
people in the scene. Body PA-MPJPE rises from 32.87\,mm at low occlusion
($<20\%$, 224 samples) to 42.11\,mm at medium ($20$--$50\%$, 136 samples) and
57.64\,mm at high occlusion ($\geq50\%$, 140 samples); the Pearson correlation
between occlusion percentage and error is 0.539.

\subsection{Limitations}
\label{sec:limitations}

CADENCE processes one tracked person per box and does not detect people or
resolve identity in multi-person scenes. The temporal refiner runs offline on
fixed-length windows, so it is not causal and cannot be used for streaming
without modification. Global orientation and limb depth remain ambiguous under
the weak-perspective camera, foreshortening, and self-occlusion, which is
consistent with the gap between CADENCE's MPJPE and PA-MPJPE on 3DPW and EMDB.
Hands are the least stable region under motion blur, hand-object contact, and
hand-hand occlusion, as the ARCTIC result shows, and the hand pathway depends
on an external 2D keypoint detector. The sign-language dynamics
metrics rely on pseudo ground truth. \Cref{sec:supp_failure_cases} summarises
failure modes.

The individual effects of kinematic conditioning, frame-dependent adaptation,
and peak-preservation supervision have not been isolated by controlled
component ablations.

\section{Conclusion}
\label{sec:conclusion}
We presented CADENCE for expressive whole-body SMPL-X recovery from monocular
video. Shared-encoder hand refinement improves finger reconstruction, while
kinematic-aware temporal filtering adapts regularisation to each joint and
frame. Experiments demonstrate improved body accuracy and temporal stability,
and sign-language evaluation shows that low temporal variation can coexist
with preservation of reference-relative hand-speed peaks. These findings
motivate evaluating hand dynamics alongside body-centric pose and acceleration
metrics. Future work will investigate causal refinement and validation using
accurately captured expressive motion.

\paragraph{Acknowledgements.}
The authors acknowledge the use of the IRIDIS X High Performance Computing
Facility and the Southampton-Wolfson AI Research Machine (SWARM) GPU cluster,
funded by the Wolfson Foundation, at the University of Southampton.

{
    \small
    \bibliographystyle{ieeenat_fullname}
    \bibliography{main}
}

\onecolumn
\appendix
\section{Implementation and Reproducibility Details}
\label{sec:supp_impl_training}

This section lists which components are frozen and trained, the supervision
used at each training stage, and the evaluation protocols.

\subsection{Architecture and Model Complexity}
\label{sec:supp_model_components}

The shared encoder serves full-person and hand crops. The pretrained weights
are frozen, with hand-specific adapters enabled during hand encoding.

\paragraph{Parameter accounting.}
\Cref{tab:supp_param_accounting} reports the parameter count for each component.
The frozen C-RADIOv4-SO400M encoder contains $472.35$M parameters and is not
updated during the main training stages. The per-frame SMPL-X estimator contains
$51.8$M trainable parameters, including the structured decoder, prediction
heads, spatial feedback modules, and lightweight adaptation parameters where
enabled. The high-resolution hand pathway adds $21.0$M trainable parameters.
The temporal refiner is lightweight, adding only $1.88$M parameters. Thus, the
per-frame and hand modules contain $72.8$M trainable parameters, and full
CADENCE including the temporal refiner contains $74.7$M trainable parameters.

\begin{table}[!htbp]
\centering
\small
\caption{
CADENCE parameter accounting. The frozen visual backbone is reported separately
from the trainable task-specific modules.
}
\label{tab:supp_param_accounting}
\setlength{\tabcolsep}{5pt}
\begin{tabular}{@{}lcc@{}}
\toprule
Component & Parameters & Trainable \\
\midrule
C-RADIOv4-SO400M visual encoder
& 472.35M & No \\
Per-frame SMPL-X estimator
& 51.8M & Yes \\
High-resolution hand pathway
& 21.0M & Yes \\
Kinematic temporal refiner
& 1.88M & Yes \\
\midrule
Per-frame + hand modules
& 72.8M & Yes \\
Full CADENCE trainable total
& 74.7M & Yes \\
\bottomrule
\end{tabular}
\end{table}

\paragraph{Per-frame estimator.}
\label{sec:supp_per_frame}

The per-frame estimator takes a person crop, resized to $384\times288$, and
predicts an initial SMPL-X parameter vector. We use C-RADIOv4-SO400M as the
shared visual encoder and extract multi-scale visual features. Early
features preserve local spatial detail, middle features capture limb
structure, and late features provide semantic context. Each feature map is
projected to the decoder width before cross-attention.

The decoder uses structured query tokens rather than a single global regression
token. Separate tokens represent camera, shape, global orientation, body joints,
left hand, right hand, jaw, and expression. Keypoint tokens represent the
projected articulated body in image space. After each decoder layer, lightweight
heads predict intermediate SMPL-X updates. The current SMPL-X estimate is
converted to joints by forward kinematics, projected to the image, and used to
update keypoint-token attention and local feature sampling in later decoder
layers. This creates an iterative spatial-feedback loop in which the model
refines both the parameter estimate and the image locations it uses for evidence.

\subsection{SMPL-X Parameterisation}
\label{sec:supp_smplx_param}

CADENCE predicts a 341-dimensional SMPL-X parameter vector for each frame:
\[
\mathbf{p}_t =
[
\boldsymbol{\theta}^{body}_t,
\boldsymbol{\theta}^{lh}_t,
\boldsymbol{\theta}^{rh}_t,
\boldsymbol{\theta}^{jaw}_t,
\boldsymbol{\psi}_t,
\boldsymbol{\theta}^{or}_t,
\boldsymbol{\beta}_t,
\mathbf{c}_t
],
\]
where body pose contains 21 body joints in 6D rotation form
($21{\times}6=126$), the left and right hands contain 15 joints each
($15{\times}6=90$ per hand), the jaw is represented by one 6D rotation,
expression has 10 dimensions, global orientation is one 6D rotation, body shape
has 10 dimensions, and the weak-perspective camera has 3 dimensions. Rotations
use the continuous 6D representation during learning and are projected to valid
rotation matrices before SMPL-X forward kinematics.

\subsection{High-Resolution Hand Pathway}
\label{sec:supp_hand_impl}

\Cref{sec:hand} describes the pathway. FreiHand and HanCo contain right hands
only; horizontal-flip augmentation exposes the shared branch to both
orientations, so no separate left-hand branch or explicit mirroring step is
needed. \Cref{fig:supp_hand_pathway} shows examples before and after the
pathway.

\subsection{Temporal Refiner Details}
\label{sec:supp_temporal_impl}

The temporal refiner is trained after the per-frame and hand components are
fixed. It operates on stored per-frame CADENCE predictions rather than raw
images. Each SMPL-X joint rotation is treated as a token over time. Tokens are first updated
with temporal self-attention and then with attention restricted to the SMPL-X
kinematic tree, allowing information to pass between each joint and its neighbours within two kinematic hops.

Each joint chooses its bandwidth per frame from a bank of four learned
symmetric kernels, initialised as the identity and Gaussians with
$\sigma{=}1,2,4$ frames. The selector's per-joint prior starts body and
global-orientation joints on the $\sigma{=}2$ kernel and the jaw, forearms, and
hands on the identity kernel, and its learned weights are zero-initialised, so
training starts from a fixed part-rate filter. Every kernel is symmetric and
non-negative with unit sum, providing centred local averaging in rotation space. The
filtered rotations are re-orthonormalised before SMPL-X forward kinematics. The
refiner has 1.88M parameters, uses width 192, two temporal layers, two tree-attention
layers, four heads, and feed-forward width 768. It uses pose only, and runs on 32-frame windows with
a hop of 16 frames and Hann-weighted overlap.

The nine-tap filters replicate boundary rotations outside each window.
Each window prediction is re-orthonormalised before overlap blending.
Overlapping predictions are averaged in the packed 6D representation using a
non-periodic Hann window with weights floored at $10^{-3}$, then divided by
accumulated weights. A final window ending at the last frame covers the tail.
Sequences of at most 32 frames use one model call. Conversion of blended 6D
rotations to matrices before forward kinematics applies Gram--Schmidt
orthonormalisation again.

\subsection{Losses and Training Schedule}
\label{sec:supp_losses}

The per-frame estimator is supervised at each decoder layer using rotation,
joint, vertex, 2D keypoint, camera, shape, and expression losses:
\[
\begin{aligned}
\mathcal{L}_{pf}
={} &
\lambda_{rot}\mathcal{L}_{rot}
+\lambda_{j3d}\mathcal{L}_{j3d}
+\lambda_{j2d}\mathcal{L}_{j2d}
+\lambda_{v}\mathcal{L}_{v} \\
&
+\lambda_{cam}\mathcal{L}_{cam}
+\lambda_{\beta}\mathcal{L}_{\beta}
+\lambda_{\psi}\mathcal{L}_{\psi}.
\end{aligned}
\]

Hand training uses geodesic finger-rotation supervision on FreiHand, HanCo,
and ReInterHand, plus HaMeR rotation distillation on COCO and UBody. Finger
supervision and distillation have weight 4, and the UBody pseudo-pose anchor
has weight 0.25. Reliable global hand orientation has weight 2 during training.

Temporal training uses reconstruction, acceleration regularisation, and
peak-speed preservation (\cref{sec:training}). Validation checkpoint acceptance
requires MPJPE regression of at most 0.2\,mm, non-increasing acceleration error,
and per-clip regression of at most 2\,mm.

\paragraph{Training stages.}
\label{sec:supp_training_stages}

CADENCE is trained in stages so that each module has a clear role.
\Cref{tab:supp_training_stages} summarises the training sequence.

\begin{table}[!htbp]
\centering
\small
\caption{
CADENCE training stages. Later stages reuse the same shared visual encoder and
preserve the previously learned whole-body prediction unless explicitly stated.
}
\label{tab:supp_training_stages}
\setlength{\tabcolsep}{4pt}
\resizebox{\textwidth}{!}{%
\begin{tabular}{@{}p{0.13\textwidth}p{0.24\textwidth}p{0.31\textwidth}p{0.22\textwidth}@{}}
\toprule
Stage & Trained modules & Supervision & Purpose \\
\midrule
Stage 1
& SMPL-X decoder and prediction heads
& Whole-body SMPL-X, 3D joints, 2D keypoints, vertices, shape, expression, camera
& Learn the base per-frame whole-body estimator \\
Stage 2
& High-resolution hand pathway
& Whole-body hand supervision and real hand-3D supervision from FreiHand/HanCo
& Improve finger articulation from hand crops \\
Stage 3
& Temporal refiner
& Stored predictions from frozen image-based components and motion supervision
& Learn adaptive filtering after image-based training \\
\bottomrule
\end{tabular}%
}
\end{table}

\paragraph{Optimisation and augmentation.}
\label{sec:supp_optim}

Training uses AdamW with mixed precision. Hand modules and hand-specific
adapters are optimised together. We apply standard image augmentations including random scale,
translation, horizontal flip, colour jitter, and synthetic occlusion where
appropriate. For video clips, we sample fixed-length windows and apply temporal
subsampling during training to improve robustness to different motion speeds.

\begin{table}[!htbp]
\centering
\small
\caption{
Optimisation summary for the training configuration.
}
\label{tab:supp_optim}
\setlength{\tabcolsep}{5pt}
\begin{tabular}{@{}ll@{}}
\toprule
Setting & Value \\
\midrule
Optimiser & AdamW \\
Precision & Mixed precision; bf16 where available \\
Decoder/head LR & $2\times10^{-4}$ \\
Whole-body training length & 300k iterations \\
Temporal window & 32 frames \\
Image augmentations & Scale, translation, flip, colour jitter, occlusion \\
Temporal augmentations & Window sampling, temporal subsampling \\
\bottomrule
\end{tabular}
\end{table}

\subsection{Training Datasets}
\label{sec:supp_training_data}

CADENCE is trained from a mixture of whole-body SMPL-X data, in-the-wild video
data, 2D keypoint supervision, and real hand-3D data. Whole-body datasets provide
the main SMPL-X supervision. COCO-WholeBody and MPII provide additional 2D
keypoint supervision. FreiHand, HanCo, and ReInterHand provide real hand-3D supervision. The temporal refiner is trained on clips from
available video sources using stored per-frame CADENCE predictions.

\begin{table}[!htbp]
\centering
\small
\caption{
Training data used by CADENCE. The table reports per-dataset image/frame counts,
modality, and the role each source plays in supervision.
}
\label{tab:supp_training_data}
\setlength{\tabcolsep}{4pt}
\resizebox{\textwidth}{!}{%
\begin{tabular}{@{}lccccl@{}}
\toprule
Dataset & Images / frames & Video & SMPL-X / 3D & 2D keypoints & Usage \\
\midrule
AGORA~\cite{Patel:CVPR:2021}
& 114{,}486 & No & Yes & Optional & Whole-body SMPL-X supervision \\
BEDLAM~\cite{Black_CVPR_2023}
& 3{,}500{,}000 & Yes & Yes & Optional & Whole-body and video supervision \\
UBody~\cite{lin2023one}
& 542{,}000 & Yes & Yes & Yes & Whole-body and expressive motion supervision \\
Motion-X~\cite{zhang2025motion}
& 2{,}931{,}318 & Yes & Yes & Optional & Motion and temporal supervision \\
3DPW~\cite{vonmarcard2018}
& 22{,}735 & Yes & Body 3D & Optional & In-the-wild body/video supervision \\
COCO-WholeBody~\cite{jin2020whole}
& 118{,}287 & No & No & Yes & 2D whole-body keypoint supervision \\
MPII~\cite{andriluka14cvpr}
& 28{,}880 & No & No & Yes & Additional 2D keypoint supervision \\
FreiHand~\cite{freihand2019}
& 130{,}240 & No & Hand 3D & Optional & Hand-pose supervision \\
HanCo~\cite{hanco2021}
& 860{,}304 & No & Hand 3D & Optional & Hand-pose supervision \\
ReInterHand~\cite{moon2023reinterhand}
& -- & Yes & Hand 3D & Yes & Hand-pose supervision \\
Public pseudo-labelled mixture
& 500{,}000 & Mixed & Pseudo & Yes & COCO, MPII, InstaVariety \\
\bottomrule
\end{tabular}%
}
\end{table}

\paragraph{Sampling.}
During per-frame training, batches are sampled from a mixture of whole-body 3D
datasets and 2D keypoint datasets. Loss masks ensure that each sample only
contributes to losses for which annotations are available. During hand-pathway
training, whole-body samples and canonical hand-only samples are mixed so that
the hand branch learns both body-context hand refinement and detailed real hand
articulation. During temporal training, the image encoder and per-frame decoder
are frozen and the temporal refiner is trained from stored predictions, ensuring
that controlled studies change only the temporal processing stage.

\subsection{Evaluation Protocols}
\label{sec:supp_eval_protocols}

EHF is evaluated with
Procrustes-aligned full-body mesh error and regional hand/face metrics. AGORA
and UBody use whole-body mesh metrics with full-body and regional breakdowns.
3DPW uses the standard 14-joint body protocol, and EMDB uses the 24-joint body
protocol. Acceleration error is computed from the difference of predicted and
ground-truth central second differences over valid consecutive frames without frame-rate scaling.

CADENCE predicts whole-body SMPL-X, but 3DPW and EMDB evaluate only the body
quantities specified by their protocols. These protocols evaluate relative body reconstruction and exclude hand
accuracy and absolute global placement.

\section{Extended Benchmark Results}
\label{sec:supp_extended_benchmarks}

This section provides additional comparisons with published methods on 3DPW.

\subsection{3DPW Extended Results}
\label{sec:supp_3dpw_extended}

3DPW is a body-centric benchmark. Although CADENCE predicts full SMPL-X
parameters, evaluation follows the body protocol used by prior work.
\Cref{tab:supp_3dpw_extended} groups methods by representation to make the
comparison clearer: whole-body SMPL-X methods, body-focused per-frame or fitting
methods, and body-only temporal methods. Acceleration errors are listed in
\cref{tab:body_compact}.

\begin{table}[!htbp]
\centering
\small
\caption{
Extended 3DPW~\cite{vonmarcard2018} body evaluation in millimetres. CADENCE
predicts whole-body SMPL-X but is evaluated with the standard 3DPW body
protocol. Dashes indicate values not reported by the corresponding method.
}
\label{tab:supp_3dpw_extended}
\setlength{\tabcolsep}{4pt}
\begin{tabular}{@{}llccc@{}}
\toprule
Method & Repr.
& PA-MPJPE $\downarrow$
& MPJPE $\downarrow$
& PVE $\downarrow$ \\
\midrule
\multicolumn{5}{@{}l}{\textit{Whole-body SMPL-X methods}} \\
\quad ExPose~\cite{choutas2020expose}
& SMPL-X & 60.7 & 93.4 & -- \\
\quad PyMAF-X~\cite{pymafx2023}
& SMPL-X & 45.3 & 74.2 & -- \\
\quad SMPLer-X-H~\cite{cai2023smplerx}
& SMPL-X & 46.6 & 76.7 & 91.8 \\
\quad PEAR~\cite{wu2026pearpixelalignedexpressivehuman}
& SMPL-X & 45.3 & 71.7 & -- \\
\quad Multi-HMR~\cite{multi-hmr2024}
& SMPL-X & 41.7 & 61.4 & -- \\
\quad DanceHMR~\cite{shen2026dancehmr}
& SMPL-X video & 35.2 & \textbf{53.6} & 65.1 \\
\rowcolor{green!10}
\quad \textbf{CADENCE}
& SMPL-X video & 40.2 & 58.1 & \textbf{62.6} \\
\midrule
\multicolumn{5}{@{}l}{\textit{Body-focused per-frame or fitting methods}} \\
\quad CameraHMR~\cite{patel2024camerahmr}
& SMPL & 35.1 & 56.0 & 65.9 \\
\quad PromptHMR~\cite{wang2025prompthmr}
& SMPL-X & 36.1 & 58.7 & 69.4 \\
\quad NLF-L+fit~\cite{sarandi2024nlf}
& SMPL & \underline{33.6} & 54.9 & 63.7 \\
\quad SAM 3D Body~\cite{sam3dbody}
& Body & \textbf{33.2} & 54.8 & \underline{63.6} \\
\midrule
\multicolumn{5}{@{}l}{\textit{Body-only temporal methods}} \\
\quad TCMR~\cite{choi2020beyond}
& SMPL video & 55.8 & 95.0 & -- \\
\quad MPS-Net~\cite{WeiLin2022mpsnet}
& SMPL video & 52.1 & 84.3 & -- \\
\quad WHAM~\cite{wham:cvpr:2024}
& SMPL video & 35.9 & 57.8 & 68.7 \\
\quad TRAM~\cite{wang2024tram}
& SMPL video & 35.6 & 59.3 & 69.6 \\
\quad GVHMR~\cite{shen2024gvhmr}
& SMPL video & 36.2 & 55.6 & 67.2 \\
\quad GENMO~\cite{li2025genmo}
& SMPL video & 34.6 & \underline{53.9} & 65.8 \\
\bottomrule
\end{tabular}
\end{table}

\section{Hand Refinement Analysis}
\label{sec:supp_hand_ablation}

All variants in this section use the same per-frame whole-body estimator, shared
encoder, and EHF protocol, and change only the hand pathway. Because the pathway
updates only finger rotations, body metrics are identical across variants.

\Cref{tab:supp_hand_supervision} compares design choices under an identical
single-stage training budget, with the CADENCE hand pathway from \cref{tab:hand_ablation}
as a separately trained reference with a different training budget. Among the listed configurations, sampling late-layer
crop features gives 12.89\,mm hand P-MVE and 8.72\,mm hand P-MPJPE, against
14.93 and 10.07\,mm with early-layer features only. Removing the zero-initialised residual or predicting the full hand pose
directly changes either error by less than 0.5\,mm.

\begin{table}[!htbp]
\centering
\small
\caption{Hand pathway design diagnostics on EHF (mm). Rows below the rule are
trained under an identical single-stage budget. The complete CADENCE
pathway is reported as a separately trained reference.}
\label{tab:supp_hand_supervision}
\setlength{\tabcolsep}{4pt}
\begin{tabular}{@{}lcc@{}}
\toprule
Variant & Hand P-MVE $\downarrow$ & Hand P-MPJPE $\downarrow$ \\
\midrule
CADENCE hand pathway (\cref{tab:hand_ablation}) & \textbf{12.2} & \textbf{8.3} \\
\midrule
Late-layer crop features & 12.89 & 8.72 \\
Early-layer crop features only & 14.93 & 10.07 \\
No zero-initialised residual & 12.45 & 8.47 \\
Predict full hand pose directly & 12.85 & 8.69 \\
\bottomrule
\end{tabular}
\end{table}

\subsection{Evaluation on Expressive Hand Articulation}
\label{sec:supp_hand_perfinger}

\Cref{tab:supp_finger} reports per-finger wrist-aligned PA-MPJPE on an internal
UBody sign-language test split with 18{,}130 samples, chosen because signing
contains frequent and varied hand articulation. Errors are largest at the
fingertips, especially of the thumb and index finger, where opposition,
self-occlusion, and fast motion make articulation hardest. CADENCE has lower
error than SMPLer-X-H~\cite{cai2023smplerx} and Multi-HMR-L~\cite{multi-hmr2024}
on every finger, with hand PA-MPVPE of 6.58\,mm against 7.63 and 7.93\,mm, and
the fraction of joints within 5\,mm (F@5\,mm) is 7 to 9 points higher.

\begin{table}[!htbp]
\centering
\caption{
Per-finger wrist-aligned PA-MPJPE (mm) on the UBody sign-language internal test
split (18{,}130 samples). Values are averaged across left and right hands.
Best results are \textbf{bold}.
}
\label{tab:supp_finger}
\footnotesize
\setlength{\tabcolsep}{2.5pt}
\begin{tabular}{@{}l cccc cccc cccc@{}}
\toprule
& \multicolumn{4}{c}{CADENCE} & \multicolumn{4}{c}{SMPLer-X-H} & \multicolumn{4}{c}{Multi-HMR-L} \\
\cmidrule(lr){2-5} \cmidrule(lr){6-9} \cmidrule(lr){10-13}
Finger & MCP & PIP & Tip & Avg & MCP & PIP & Tip & Avg & MCP & PIP & Tip & Avg \\
\midrule
Thumb  & \textbf{4.45} & \textbf{3.76} & \textbf{8.21} & \textbf{5.47} & 5.23 & 4.31 & 9.48 & 6.34 & 5.13 & 4.46 & 9.86 & 6.48 \\
Index  & \textbf{3.06} & \textbf{4.35} & \textbf{8.02} & \textbf{5.15} & 3.77 & 4.88 & 9.40 & 6.02 & 3.52 & 5.38 & 10.97 & 6.62 \\
Middle & \textbf{2.79} & \textbf{3.51} & \textbf{6.65} & \textbf{4.32} & 3.37 & 3.94 & 7.54 & 4.95 & 3.34 & 4.23 & 7.58 & 5.05 \\
Ring   & \textbf{2.49} & \textbf{2.34} & \textbf{5.30} & \textbf{3.38} & 2.96 & 2.68 & 6.26 & 3.96 & 2.95 & 2.94 & 6.30 & 4.06 \\
Pinky  & \textbf{2.79} & \textbf{3.27} & \textbf{6.13} & \textbf{4.06} & 3.35 & 3.50 & 6.36 & 4.40 & 3.29 & 3.78 & 6.94 & 4.67 \\
\midrule
Avg    & \textbf{3.12} & \textbf{3.45} & \textbf{6.86} & \textbf{4.48} & 3.74 & 3.86 & 7.81 & 5.13 & 3.65 & 4.16 & 8.33 & 5.38 \\
\midrule
\multicolumn{1}{@{}l}{Hand PA-MPVPE} & \multicolumn{4}{c}{\textbf{6.58}} & \multicolumn{4}{c}{7.63} & \multicolumn{4}{c}{7.93} \\
\midrule
\multicolumn{1}{@{}l}{F@5\,mm (\%$\uparrow$)} & \multicolumn{4}{c}{\textbf{71.11}} & \multicolumn{4}{c}{64.16} & \multicolumn{4}{c}{61.88} \\
\multicolumn{1}{@{}l}{F@15\,mm (\%$\uparrow$)} & \multicolumn{4}{c}{\textbf{96.85}} & \multicolumn{4}{c}{95.91} & \multicolumn{4}{c}{95.54} \\
\bottomrule
\end{tabular}
\end{table}

\section{Temporal Refinement Analysis}
\label{sec:supp_temporal}

This section complements \cref{sec:temporal_results}. All variants start from
the same stored per-frame CADENCE predictions, evaluation clips, and metric
implementation, and change only the temporal stage.

\subsection{Shared and Joint-Dependent Fixed Filtering}
\label{sec:supp_temporal_components}

\Cref{tab:supp_temporal_components} uses matched 32-frame clip subsets of 3DPW
and SignAvatars/How2Sign, with identical stored inputs within each subset. The comparison evaluates fixed shared-bandwidth and joint-dependent filters.
The subsets differ from the main evaluation, so values are interpreted within
this table.

\begin{table}[!htbp]
\centering
\small
\caption{Fixed temporal filtering on matched 32-frame evaluation subsets.
All variants within a dataset use identical stored per-frame inputs.
These configurations compare shared and joint-dependent fixed bandwidths.}
\label{tab:supp_temporal_components}
\setlength{\tabcolsep}{4pt}
\begin{tabular}{@{}lcccccc@{}}
\toprule
Variant
& \multicolumn{3}{c}{3DPW diagnostic subset}
& \multicolumn{3}{c}{SignAvatars/How2Sign diagnostic subset} \\
\cmidrule(lr){2-4}\cmidrule(lr){5-7}
& PA-MPJPE $\downarrow$
& MPJPE $\downarrow$
& Accel $\downarrow$
& Jitter $\downarrow$
& Peak-R $\uparrow$
& Peak-A (ideal 1) \\
\midrule
CADENCE per-frame
& 42.07 & 61.74 & 16.96
& 3.47 & 0.93 & 1.15 \\
\midrule
Single-rate temporal filter
& 41.67 & 61.20 & 5.72
& 0.63 & 0.63 & 0.55 \\
Multi-rate temporal filter
& 41.67 & 61.20 & 5.72
& 0.63 & 0.76 & 0.63 \\
\bottomrule
\end{tabular}
\end{table}

The single-rate and multi-rate filters have identical body accuracy,
acceleration error, and jitter, but the multi-rate filter keeps more of the hand
motion: Peak-R rises from 0.63 to 0.76 and Peak-A from 0.55 to 0.63. At comparable temporal smoothness, joint-dependent filtering improves
hand-velocity peak preservation relative to uniform filtering.

\subsection{Body Reconstruction Robustness}
\label{sec:supp_temporal_robustness}

Temporal filters can be sensitive to frame rate and missing observations. We
therefore evaluate CADENCE temporal refinement under simple test-time
degradations: temporal downsampling and random frame dropping. In each case, the
same per-frame model is used and only the temporal input sequence is degraded.
For downsampling, we evaluate on the retained frames. For missing frames, dropped
frames are masked during temporal refinement and metrics are computed on frames
with available ground truth.

\begin{table}[!htbp]
\centering
\small
\caption{
Body reconstruction robustness under low frame rate and missing frames.
Test-time degradations are applied to the temporal input sequence only. Body accuracy is reported on 3DPW. Acceleration uses unscaled second
differences on retained frames, whose spacing changes under subsampling.
}
\label{tab:supp_temporal_robustness}
\setlength{\tabcolsep}{4pt}
\begin{tabular}{@{}lccc@{}}
\toprule
Setting & PA-MPJPE $\downarrow$ & MPJPE $\downarrow$ & Accel $\downarrow$ \\
\midrule
Original frame rate & 40.2 & 58.1 & 5.4 \\
Half frame rate & 42.1 & 59.5 & 12.3 \\
One-third frame rate & 45.0 & 61.9 & 21.6 \\
Random drop 10\% & 40.4 & 58.2 & 7.1 \\
Random drop 25\% & 40.8 & 58.6 & 10.0 \\
\bottomrule
\end{tabular}
\end{table}

Accuracy degrades progressively under temporal subsampling. Random frame
dropping is handled robustly: PA-MPJPE rises from $40.2$ to $40.8$\,mm when a
quarter of frames are removed. Temporal downsampling is more demanding because
the inter-frame interval grows, which both widens the motion gap the refiner must
bridge and increases the measured finite-difference acceleration. Even so,
accuracy degrades smoothly, with PA-MPJPE changing from $40.2$ to $45.0$\,mm
from full to one-third frame rate.

\subsection{Learned Temporal Kernels}
\begin{figure}[!htbp]
\centering
\includegraphics[width=0.88\linewidth]{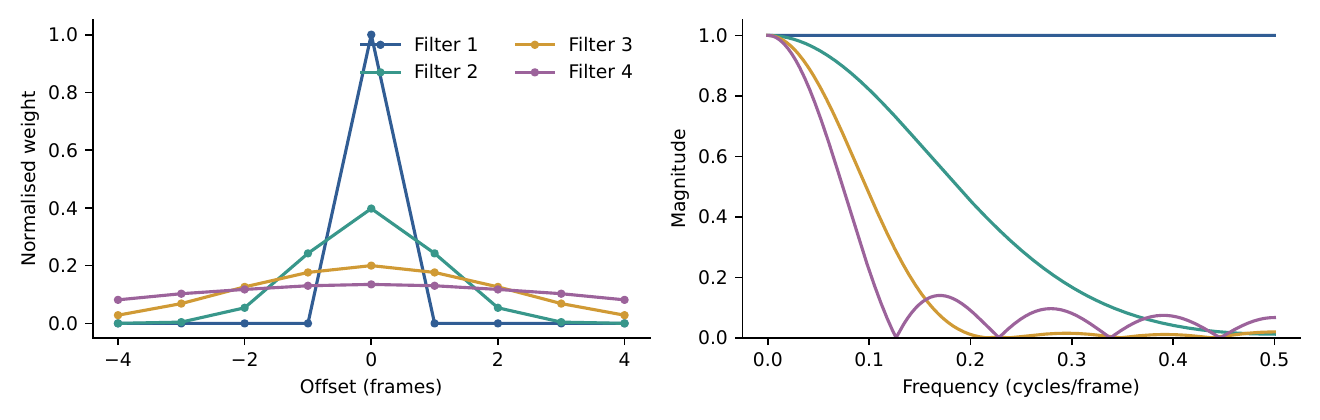}
\caption{Learned kernels and frequency-response magnitudes extracted from the
trained CADENCE checkpoint. Filter indices denote initial bank slots.
Symmetry and normalisation constrain local averaging, while the learned
frequency responses need not be monotonically ordered.}
\label{fig:learned_filters}
\end{figure}

\Cref{fig:learned_filters} shows the four kernels from the trained checkpoint.
The first remains close to the identity and has an approximately flat response.
The other kernels distribute weight across neighbouring frames, with stronger
high-frequency attenuation and, for the broadest kernel, response sidelobes.
Maximum absolute tap changes from initialisation are approximately
$8\times10^{-10}$, 0.0013, 0.0040, and 0.0012, respectively. The bank thus
retains distinct responses with modest changes in individual weights.
These are fixed-kernel frequency responses. Joint- and frame-dependent mixing
and rotation projection give the complete refiner no single global response.

\subsection{Rotation Interpolation and Temporal Filtering Details}
\label{sec:supp_rotation_filtering}

CADENCE applies temporal processing to articulated SMPL-X rotations rather than
to unconstrained 3D joint coordinates. Rotations are represented using the
continuous 6D representation during learning, converted to valid rotation
matrices before forward kinematics, and re-orthonormalised after temporal
updates. The temporal model does not require linear interpolation of
reduced-rate 6D rotations back to the full frame rate; the per-joint temporal
filters operate over the full input window with different effective bandwidths
for different joint groups.

For diagnostic variants that explicitly resample time, interpolation is applied
on valid rotations rather than on arbitrary unnormalised 6D vectors. In
practice, we convert 6D rotations to rotation matrices, interpolate rotations
using a rotation-aware procedure, and then convert back to the representation
used by the model. This avoids treating the 6D representation as an unconstrained
Euclidean signal and prevents invalid rotations from entering the SMPL-X forward
kinematics.

\section{Computational Efficiency}
\label{sec:supp_runtime}

\Cref{tab:supp_runtime_accounting} reports throughput for the main inference configurations and the dominant runtime components. All measurements are obtained on one A100-80GB GPU using bf16 inference. Batch-1 throughput measures low-latency inference, while batch-32 throughput measures batched throughput.

The base per-frame estimator processes the full-person crop and predicts
whole-body SMPL-X parameters. Enabling the high-resolution hand pathway adds two
hand-crop encoder passes, improving finger articulation but reducing throughput.
The temporal refiner is inexpensive relative to image encoding, running at
approximately $6{,}900$ FPS for a 32-frame window.

\begin{table}[!htbp]
\centering
\small
\caption{
A100-80GB bf16 runtime and memory measurements. The visual encoder dominates
runtime, the high-resolution hand pathway trades throughput for improved finger
articulation, and the temporal refiner adds negligible overhead.
}
\label{tab:supp_runtime_accounting}
\setlength{\tabcolsep}{4pt}
\begin{tabular}{@{}lccc@{}}
\toprule
Configuration & Batch 1 FPS & Batch 32 FPS & Notes \\
\midrule
Visual encoder only
& 62.8 & 367.0
& Shared C-RADIOv4 encoder \\
Base per-frame estimator
& 27.9 & 307.6
& Full-person crop; encoder + decoder + heads \\
Full CADENCE
& 13.1 & 71.5
& Per-frame estimator + hand pathway + temporal refiner \\
Temporal refiner only
& $\sim$6{,}900 & $\sim$6{,}900
& 32-frame window; negligible overhead \\
\midrule
Peak GPU memory
& \multicolumn{2}{c}{3.73 GiB}
& Batch 32, bf16 \\
\bottomrule
\end{tabular}
\end{table}

Visual encoding and high-resolution hand processing dominate runtime.
The hand pathway trades throughput for improved finger articulation by
reusing the same encoder on two additional crops.
Applications that require maximum throughput can run the per-frame
estimator with the temporal refiner and skip the hand pathway.

\section{Additional Qualitative Results}
\label{sec:supp_qual_failure}

This section describes layer-by-layer decoder refinement, a comparison with
whole-body baselines, temporal examples, and failure cases. The figures illustrate the corresponding spatial and temporal behaviours.

\paragraph{Videos.} Video results, including continuous two-handed
sign-language sequences, are on the project page,
\url{https://keniel123.github.io/cadence}.

\subsection{Failure Cases}
\label{sec:supp_failure_cases}

Monocular ambiguity and occlusion remain limitations of CADENCE
(\cref{sec:limitations}). The ARCTIC analysis measures increased error under
hand-object interaction, and the AGORA occlusion analysis quantifies reduced
body accuracy as visibility decreases.

\subsection{Whole-Body Qualitative Results}
\label{sec:supp_qualitative_wholebody}

\begin{figure}[!htbp]
\centering
\includegraphics[width=\textwidth]{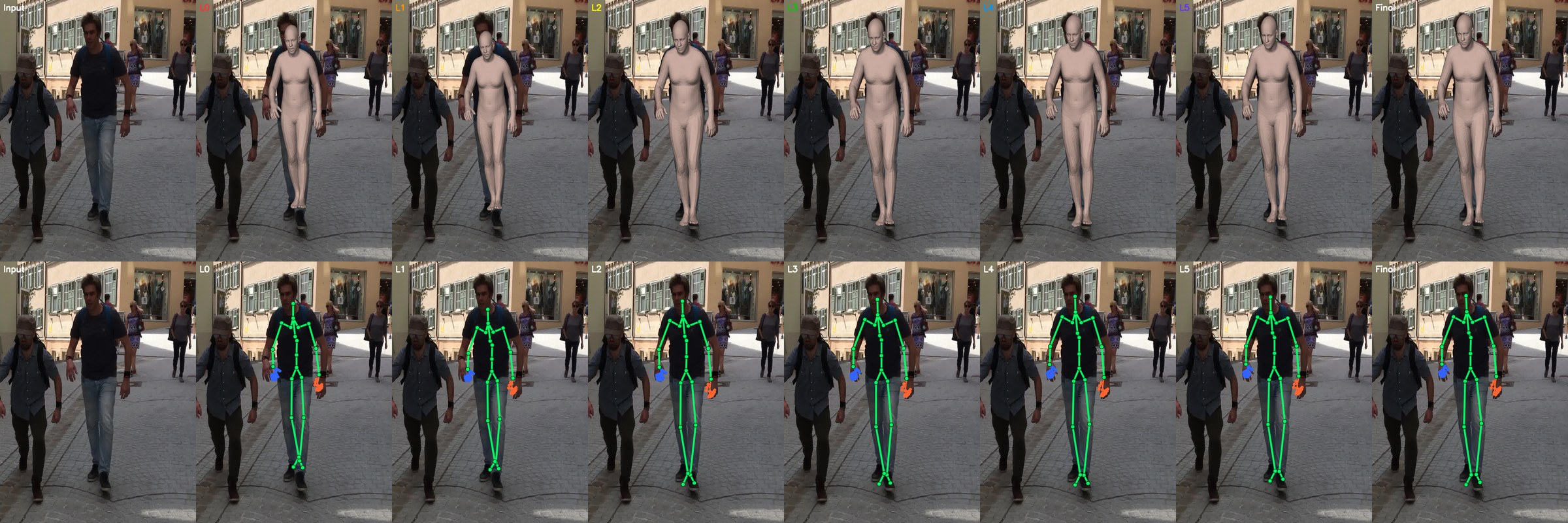}\\[3pt]
\includegraphics[width=\textwidth]{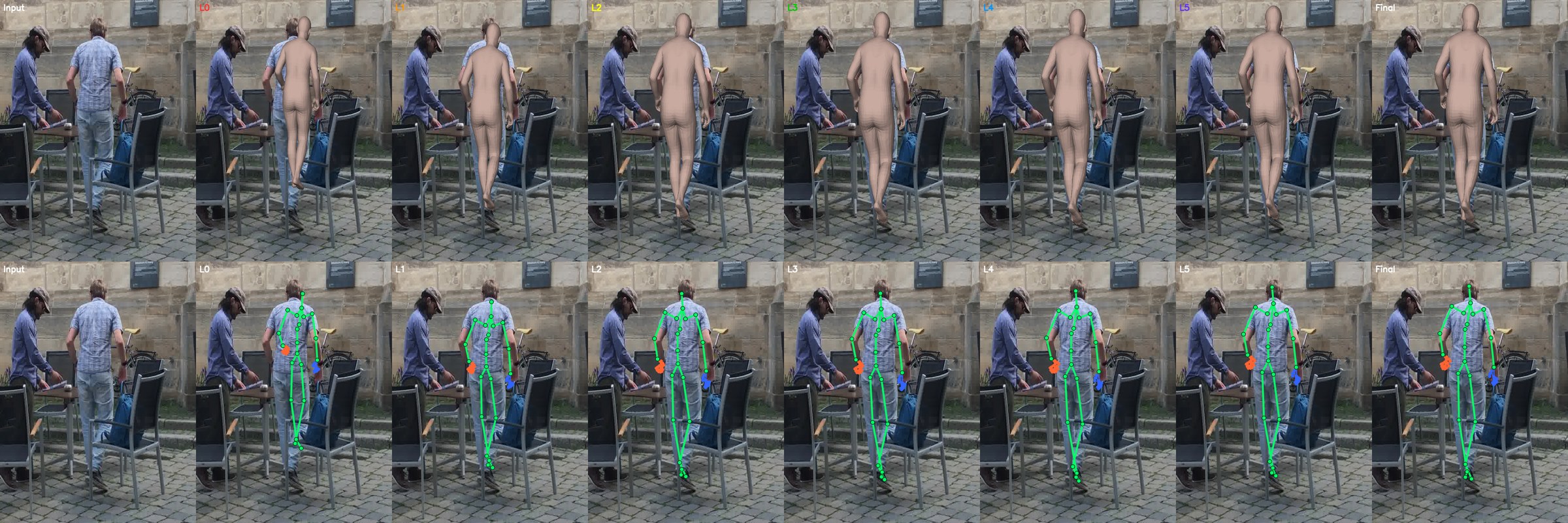}\\[3pt]
\includegraphics[width=\textwidth]{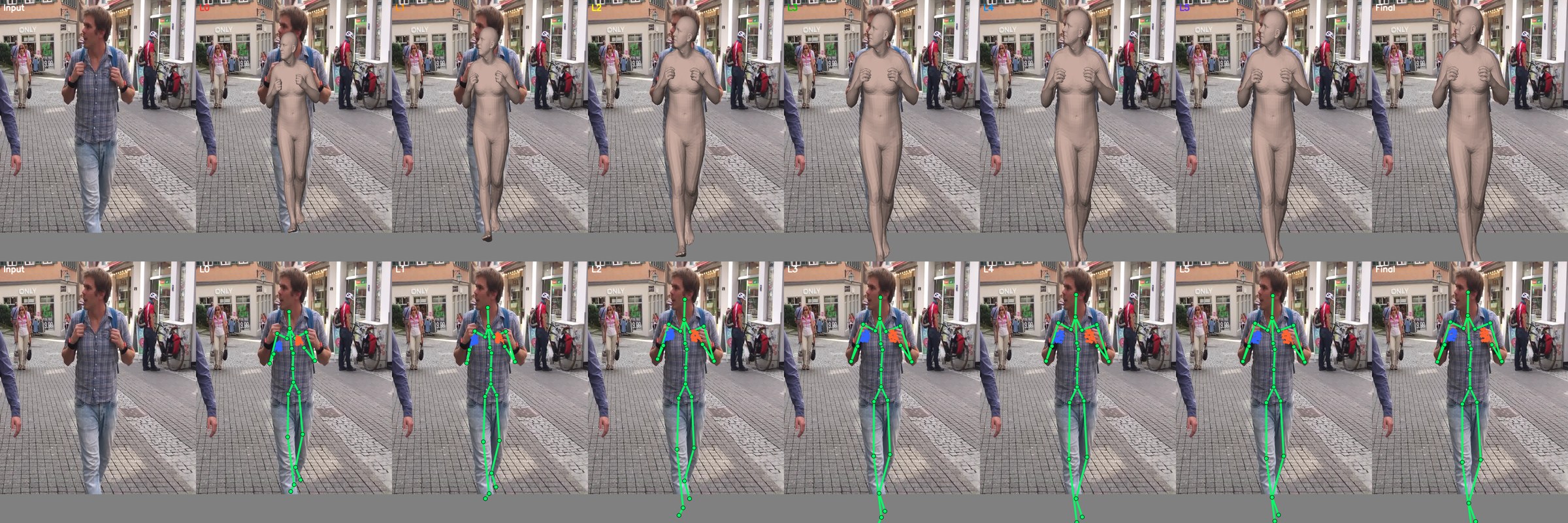}\\[3pt]
\caption{Whole-body qualitative results with layer-by-layer refinement on
3DPW~\cite{vonmarcard2018} test sequences. For each example, the top row shows
the SMPL-X mesh overlay across decoder layers
(Input$\to$L0$\to$L1$\to$L2$\to$L3$\to$L4$\to$L5$\to$Final), and the bottom row
shows the corresponding skeleton keypoints. The iterative decoder progressively
improves mesh alignment and joint localisation across layers. Please zoom in for
details.\label{fig:supp_qual_wholebody}}
\end{figure}

\Cref{fig:supp_qual_wholebody} shows additional CADENCE reconstructions on
diverse 3DPW sequences, visualised as a layer-by-layer refinement progression.
The structured decoder starts from a coarse initial estimate and progressively
updates the mesh and joint locations across its six layers using iterative
spatial feedback. The examples include walking, running, stair climbing, and
seated poses with partial occlusion. They illustrate how the decoder refines
global body structure and projected joint alignment over successive layers.

\subsection{Multi-Baseline Qualitative Comparison on Expressive Scenes}
\label{sec:supp_qual_multibaseline}

\begin{figure}[!htbp]
\centering
\includegraphics[height=0.80\textheight]{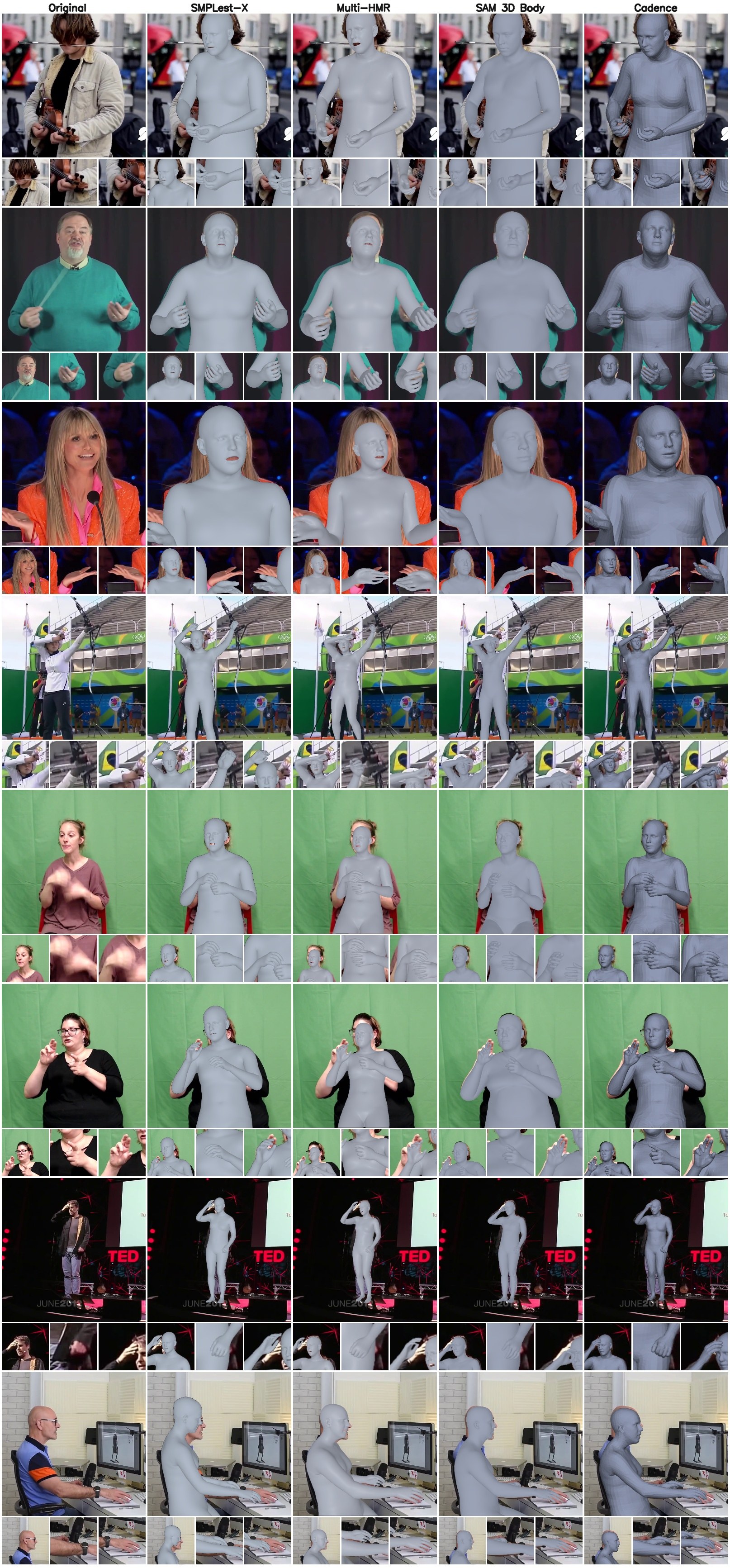}
\caption{Qualitative comparison on expressive UBody~\cite{lin2023one} scenes. Columns
show the input frame, SMPLer-X~\cite{cai2023smplerx},
Multi-HMR~\cite{multi-hmr2024}, SAM~3D~Body~\cite{sam3dbody}, and CADENCE with
the hand pathway. Each method is shown with a full-person overlay and a zoom
strip for the face and both hands using shared crop regions. This figure is
qualitative context rather than a hand-accuracy comparison. SAM~3D~Body is
shown with a non-SMPL-X topology. Please zoom in for details.\label{fig:supp_qual_multibaseline}}
\end{figure}

\Cref{fig:supp_qual_multibaseline} compares CADENCE qualitatively with recent
whole-body or whole-human recovery methods on expressive UBody scenes. We compare
against SMPLer-X~\cite{cai2023smplerx}, Multi-HMR~\cite{multi-hmr2024}, and
SAM~3D~Body~\cite{sam3dbody}. To make the visual comparison controlled, all
methods are run in single-person mode using the same detected person box where
possible. Each panel shows the full-person overlay and a zoom strip containing
the face and both hands cropped from the same image regions.

The methods use different representations and processing assumptions in
expressive regions such as hands and face. SAM~3D~Body uses a different mesh
representation from SMPL-X. We therefore show the panels only as qualitative
context and avoid topology-specific or hand-accuracy claims. CADENCE's
high-resolution branch is finger-only and retains a whole-body SMPL-X output.

\subsection{Consecutive-Frame Visualisations Across Diverse Activities}
\label{sec:supp_qual_temporal_consistency}

\begin{figure}[!htbp]
\centering
\includegraphics[width=0.9\textwidth]{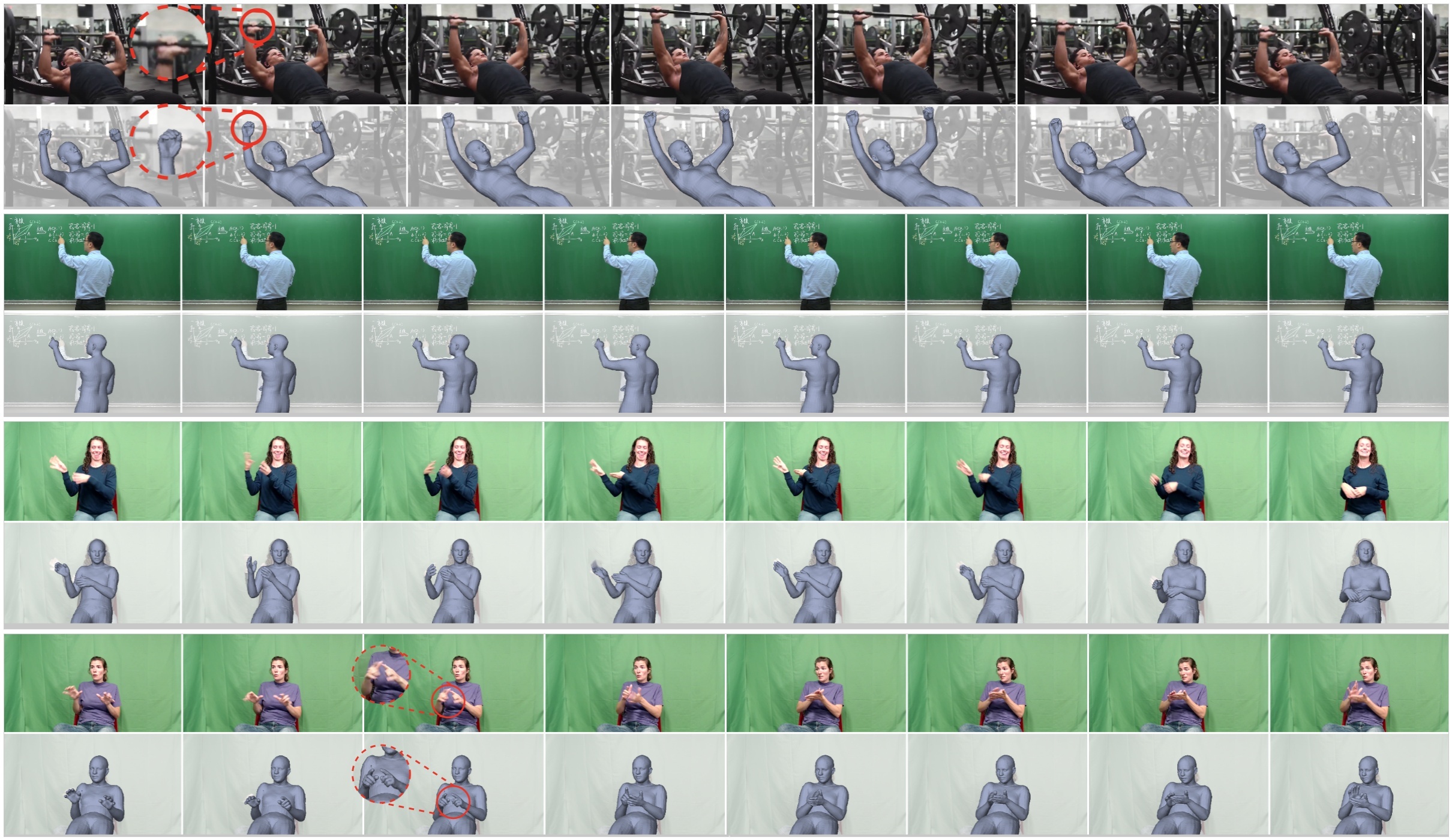}
\caption{Consecutive-frame CADENCE visualisations across diverse activities. Each clip
is shown as a pair of rows: input video frames and CADENCE SMPL-X overlays,
sampled at consecutive timesteps from left to right. The examples include
dynamic body motion, classroom interaction, and expressive sign-language
sequences. Highlighted insets mark hand regions across neighbouring frames.
Please zoom in for details.\label{fig:supp_qual_temporal_consistency}}
\end{figure}

\Cref{fig:supp_qual_temporal_consistency} shows CADENCE reconstructions on clips
spanning different activity types, including dynamic athletic motion, classroom
interaction, and expressive sign-language sequences. For each clip, consecutive
frames are shown from left to right with the recovered SMPL-X overlay. These
examples show how the recovered body and hand configurations evolve across
neighbouring frames in diverse scenes.

\end{document}